\documentclass[10pt,twocolumn,letterpaper]{article}

\usepackage{multirow}
\usepackage[pagenumbers]{cvpr}      

\usepackage[pagebackref,breaklinks,colorlinks]{hyperref}
\usepackage[hyphens]{url}
\usepackage{algorithm}
\usepackage{algorithmic}
\usepackage{amsmath}
\usepackage{cancel}

\newlength\savedwidth
\usepackage{booktabs}
\usepackage{bm}

\usepackage{physics}
\usepackage{amssymb}
\usepackage{amsmath}
\usepackage{cancel}
\usepackage{mathtools}
\mathtoolsset{showonlyrefs}

\title{\textbf{Robust Human Trajectory Prediction via
Self-Supervised Skeleton Representation
Learning}}
\author{
    Taishu Arashima$^{1,\dagger}$
    \;
    Hiroshi Kera$^{2,\dagger}$
    \;
    Kazuhiko Kawamoto$^{3,\dagger}$
    \\
    $^{\dagger}$Chiba University
    \\
    {\tt\small $^1$taishu.arashima@chiba-u.jp},
    {\tt\small $^2$kera@chiba-u.jp},
    {\tt\small $^3$kawa@faculty.chiba-u.jp}
}
\date{}

\begin{document}

\maketitle

\begin{abstract}
Human trajectory prediction plays a crucial role in applications such as autonomous navigation and
video surveillance. While recent works have explored the integration of human skeleton sequences to
complement trajectory information, skeleton data in real-world environments often suffer from missing joints caused by occlusions. These disturbances significantly degrade prediction accuracy, indicating the need for more robust skeleton representations.
We propose a robust trajectory prediction method that incorporates a self-supervised skeleton
representation model pretrained with masked autoencoding. 
Experimental results in occlusion-prone scenarios show that our method improves robustness to missing skeletal data without sacrificing prediction accuracy, and consistently outperforms baseline models in clean-to-moderate missingness regimes.
\end{abstract}

 \section{Introduction}
\label{sec:intro}
Human trajectory prediction estimates future pedestrian paths from past motion history and is a fundamental technique for applications such as path planning for autonomous robots \cite{7490340} and human tracking in surveillance systems \cite{6126296}. 
Accurate prediction requires understanding pedestrian intention, which is often difficult to infer from positions alone.
Most existing trajectory prediction methods \cite{Alahi_2016_CVPR, 8578338, Mohamed_2020_CVPR, 10.1109/TITS.2021.3069362, Salzmann2020TrajectronPP} use only past trajectories as input. As a result, these methods often struggle to capture intention-related cues such as directional changes and characteristic motion patterns. 
To address this limitation, recent studies incorporate human skeletal sequences as an auxiliary modality \cite{robots_can_see, saadatnejad2024socialtransmotion}. Because skeletal sequences directly reflect human body structure, skeletal information provides high-level motion cues that are not available from trajectories alone.
However, skeletal sequences collected in real environments are often incomplete. Joint detection errors, body orientation changes, and occlusions frequently cause partial missing joints in the input skeleton. 
Such incomplete skeletal observations disrupt the structural and temporal consistency of skeletal representations, leading to substantial performance degradation. 
This issue remains a challenge for skeleton-aided trajectory prediction methods under diverse missing patterns.
A straightforward approach is to train the downstream predictor directly on incomplete skeleton inputs so that it adapts to missingness. However, this strategy often introduces a practical trade-off: while it can improve tolerance to missing joints, it may weaken the use of informative skeleton cues under clean conditions and thus reduce prediction accuracy.

\begin{figure}[t]
    \centering
    \includegraphics[width=0.45\textwidth]{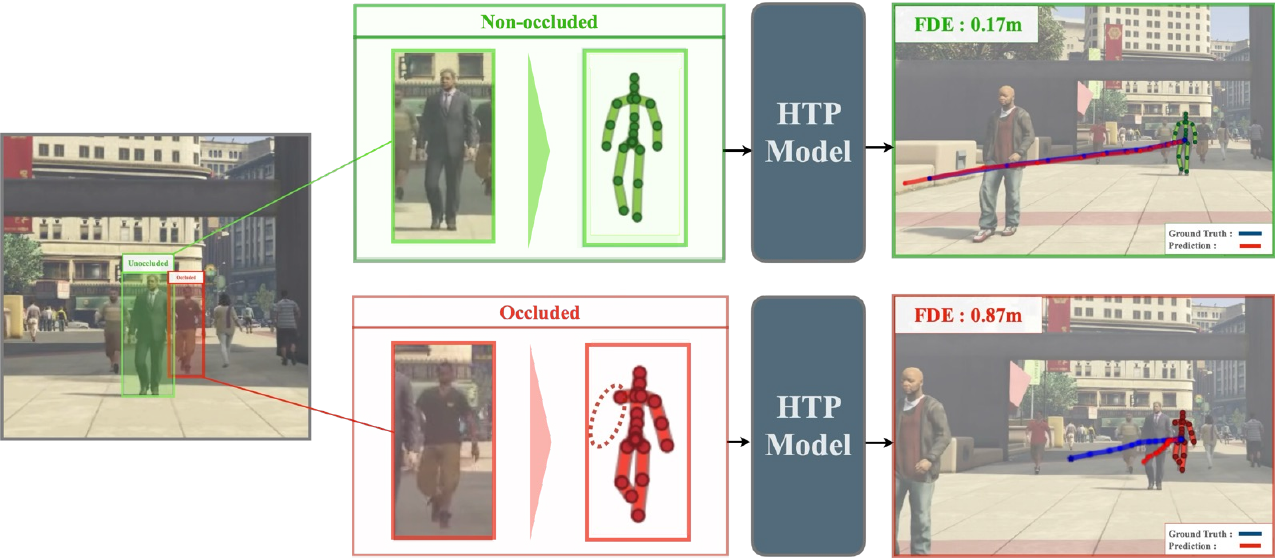}
    \caption{Impact of occlusion-induced missing skeletons on trajectory prediction accuracy.
The top row shows a clean (non-occluded) skeleton observation, while the bottom row shows an occluded observation with missing joints. Using an existing human trajectory prediction (HTP) model, prediction accuracy is high under clean inputs (FDE = 0.17 m) but degrades substantially under occluded inputs (FDE = 0.87 m), where FDE denotes the final displacement error.
}
    \label{fig:intro_overview}
\end{figure}

To address this trade-off, we propose a two-stage framework that separates self-supervised skeleton representation learning from downstream trajectory prediction. In the pretraining stage, a subset of joints is intentionally masked, and the model is trained to reconstruct the masked joint coordinates from incomplete inputs. This process enables the encoder to learn stable skeleton representations under partial observability.
The pretrained encoder is then integrated into the downstream predictor as a skeleton embedding module and fused with trajectory information to provide body-motion cues for prediction. The key idea of our method is not to make the downstream predictor itself adapt to incomplete skeletons, but to acquire robustness at the representation level during pretraining. This design aims to preserve the ability to exploit skeleton cues while reducing sensitivity to missing joints, thereby alleviating the trade-off between prediction accuracy and robustness.

For evaluation, we use the large-scale synthetic JTA dataset \cite{fabbri2018learning} and construct multiple missing-data scenarios with different corruption settings. Experimental results show that the proposed method consistently improves prediction accuracy from clean to moderate missingness conditions while remaining competitive under severe missingness. In addition, skeleton-reliance analysis and qualitative comparisons confirm that the gains do not come from weakening skeleton usage, but from preserving the usefulness of skeleton features under partial observations.

The main contributions of this paper are summarized as follows.
\begin{itemize}
\item \textbf{A two-stage framework for robust skeleton-aided trajectory prediction.}
We propose an integrated framework that explicitly separates reconstruction-based skeleton representation learning from downstream trajectory prediction, improving tolerance to missing skeleton inputs while preserving prediction performance under clean conditions.
\item \textbf{Representation-level robustness via mask-based self-supervised skeleton learning.}
We show that reconstruction-based pretraining learns skeleton representations that remain informative across diverse missing patterns, positioning mask-based self-supervision as a principled approach to robustness rather than only a task-specific performance enhancement.
\end{itemize}

 \section{Related Work}
\label{sec:related}

\medskip\noindent\textbf{Human Trajectory Prediction}
Early work on human trajectory prediction models pedestrian motion using physics-inspired formulations \cite{helbing1995social} or statistical approaches \cite{kim2011gaussian,wang2007gaussian}. These methods provide interpretable models of motion dynamics but rely on simplified assumptions that limit their ability to capture complex, nonlinear social behaviors.
With the rise of deep learning, data-driven approaches have become dominant. Recurrent neural networks (RNNs) \cite{Alahi_2016_CVPR,8578338,10.1109/TITS.2021.3069362,Salzmann2020TrajectronPP} model temporal dependencies in motion sequences, while attention-based models \cite{9412190,9577379,9710708,saadatnejad2024socialtransmotion} selectively reason over socially relevant agents and time steps. In parallel, graph-based methods \cite{9560908,Mohamed_2020_CVPR} explicitly encode inter-agent interactions and achieve strong performance in crowded scenes.
Despite architectural diversity, most of these approaches share an implicit assumption: pedestrian intent can be inferred primarily from spatial trajectories. This position-based representation often fails to capture fine-grained motion cues that reflect internal intent, especially in socially complex environments.

To address this limitation, recent work incorporates human skeletal information as an additional modality. Yagi et al.\ \cite{Yagi_2018_CVPR} extract discriminative features from skeletal sequences to improve accuracy. Salzmann et al.~\cite{robots_can_see} integrate 3D pose, head orientation, and scene context into a unified Transformer framework, demonstrating the importance of explicitly modeling body dynamics. Subsequent methods, including SocialTransMotion \cite{saadatnejad2024socialtransmotion} and its extensions \cite{gao2024multitransmotion,Gao2025SocialPoseET}, further confirm that skeletal cues provide complementary information beyond trajectories.
However, these approaches implicitly assume reliable skeletal observations. In real-world settings, pose estimates are often noisy or incomplete due to occlusions, viewpoint changes, and sensing errors. Naively incorporating such imperfect skeletal data can degrade performance and obscure the intended benefits of multimodal modeling.

\medskip\noindent\textbf{Robustness to Incomplete Skeletons}
Skeletal representations are inherently vulnerable to observation noise. Occlusions, sensor limitations, and pose estimation failures frequently lead to missing or corrupted joints, which can significantly affect downstream tasks.
Prior work addresses this issue primarily through explicit reconstruction of missing joints. Autoencoder-based methods exploit structural priors of the human body to recover plausible poses from incomplete inputs \cite{Carissimi_2018_ECCV_Workshops,10.1145/3197517.3201302}. Temporal models leverage motion continuity to infer missing information across time \cite{10.5555/3367032.3367133,10.1016/j.cag.2019.03.010}. In parallel, the pose estimation community improves robustness at the sensing stage through occlusion-aware architectures \cite{9010921}, spatio-temporal Transformers \cite{10.1145/3730436.3730463,10204481}, and diffusion-based generative models that produce multiple plausible hypotheses under severe ambiguity \cite{10377372}.

While effective for pose recovery, these approaches typically aim to reconstruct a single deterministic pose sequence. Reconstruction errors therefore propagate directly to downstream predictors. Moreover, accurate reconstruction alone does not necessarily guarantee that the resulting representation is robust or informative for prediction tasks.
In contrast, we argue that robustness should be learned at the representation level rather than enforced through reconstruction. By training models to encode partially observed skeletons into stable latent representations, downstream predictors can operate on features that are inherently robust to missing or corrupted inputs.

\medskip\noindent\textbf{Skeleton Representation Learning}
Self-supervised learning has emerged as an effective paradigm for learning skeletal representations without manual annotation \cite{9713748}. By exploiting the intrinsic spatiotemporal structure of human motion, these methods achieve strong performance across a wide range of downstream tasks.
Existing approaches fall broadly into two categories. Contrastive methods enforce consistency across augmented views of the same motion sequence \cite{RAO202190,10.1145/3474085.3475307,guo2022aimclr,10656009, mehraban2025stars}, encouraging invariance to perturbations. Generative methods reconstruct masked or corrupted skeletal inputs, learning latent representations that capture spatial and temporal dependencies \cite{Zheng_Wen_Liu_Long_Dai_Gong_2018,su2020predict,10222534,10376785}.

Despite strong empirical results, most prior work evaluates representations primarily in terms of downstream accuracy under clean or mildly perturbed conditions. This focus implicitly assumes reliable skeletal observations and leaves robustness under realistic partial observability underexplored.
In this work, we reinterpret masked modeling not merely as a reconstruction objective but as a mechanism for learning uncertainty-aware and robust skeletal representations. By explicitly training under partial observability, the learned representations generalize more reliably to real-world scenarios characterized by missing skeletal data.

 \section{Proposed Method} \label{method}

We propose a trajectory prediction framework that incorporates 3D skeleton information under partial observability.
The method pretrains a spatio-temporal skeleton encoder via masked joint reconstruction and replaces the linear pose embedding in Social-TransMotion with the pretrained encoder output.

\subsection{Self-Supervised Skeleton Representation Learning} \label{sec:preexp}
We design the skeleton encoder with robustness to missing and corrupted observations.
We pretrain the encoder using a reconstruction-based self-supervised learning framework.
During pretraining, a subset of joints is masked from the input skeleton sequence, so that the encoder learns to infer missing joint information from the remaining visible joints and temporal context.

\subsubsection{Skeleton Representation}
We adopt an asymmetric encoder-decoder architecture for self-supervised skeleton representation learning.
The encoder maps a masked skeleton sequence to joint-level latent features, while a lightweight decoder reconstructs masked joint coordinates from the latent representation.

We represent the human skeleton as a graph composed of joints and their physical connections.
Let $\mathbf{x}_{t,i} \in \mathbb{R}^C$ denote the coordinate of joint $i$ at time step $t$.
The skeleton at time $t$ is defined as
\[
\mathbf{X_t} =
\begin{bmatrix}
\mathbf{x}_{t,1}^\top \\
\mathbf{x}_{t,2}^\top \\
\vdots \\
\mathbf{x}_{t,N}^\top
\end{bmatrix}
\in \mathbb{R}^{N \times C}.
\]
where $N$ is the number of joints and $C$ is the coordinate dimensionality.
A skeleton sequence consisting of $T$ frames is defined as
\[
\mathbf{S} = (X_1, X_2, \ldots, X_T)
\in \mathbb{R}^{T \times N \times C}.
\]
The skeleton structure is defined by a joint set $V=\{v_1,\ldots,v_N\}$ and an adjacency matrix
$A \in \{0,1\}^{N \times N}$, 
where $A_{ij}=1$ indicates a physical connection between joints $i$ and $j$.

\begin{figure}[t]
    \centering
    \includegraphics[width=0.48\textwidth]{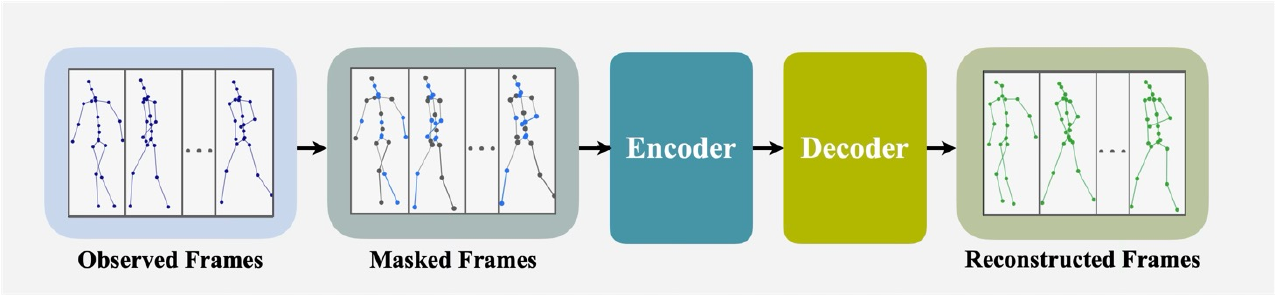}
    \caption{Overview of the self-supervised skeleton learning framework.
    Random masks are applied to the input skeleton sequence, and a spatio-temporal encoder extracts latent representations from the remaining visible joints.
    A decoder then reconstructs the original skeleton sequence.
    Through this process, the model learns structural representations invariant to missing data and noise.}
    \label{fig:pretrain_methods}
\end{figure}

\subsubsection{Architecture}
The encoder $G_E$ is implemented as an $L$-layer ST-GCN\cite{Yan_Xiong_Lin_2018},
which jointly models spatial dependencies across frames.
Given a masked skeleton sequence $\tilde{\mathbf{S}}$, the encoder produces a latent representation by leveraging the remaining visible joints and the underlying spatio-temporal structure:
\begin{equation}
\mathbf{H} = G_E(\mathbf{A}, \tilde{\mathbf{S}}), \quad \mathbf{H} \in \mathbb{R}^{T \times N \times D}.
\end{equation}
where $D$ is the feature dimension.

The decoder $G_D$ maps the latent representation back to joint coordinates:
\begin{equation}
\mathbf{Y} = G_D(\mathbf{A}, \mathbf{H}), \quad \mathbf{Y} \in \mathbb{R}^{T \times N \times C}.
\end{equation}
Following the asymmetric design of masked autoencoders, the decoder is implemented as a lightweight MLP.
The lightweight decoder prevents compensation at the decoding stage.
This design encourages the encoder to capture dependencies that are transferable to downstream tasks.
The encoder $G_E$ is retained for downstream trajectory prediction.

\begin{figure}[t]
  \centering
  \begin{subfigure}[t]{0.15\textwidth}
    \centering
    \includegraphics[width=\linewidth]{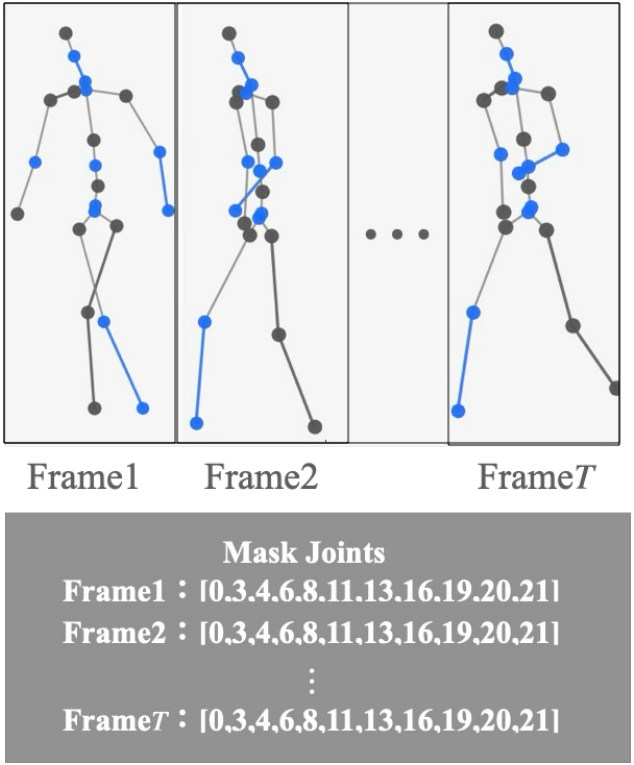}
    \caption{Temporally Consistent mask
    }
    \label{fig:mask_pattern_a}
  \end{subfigure}\hfill
  \begin{subfigure}[t]{0.15\textwidth}
    \centering
    \includegraphics[width=\linewidth]{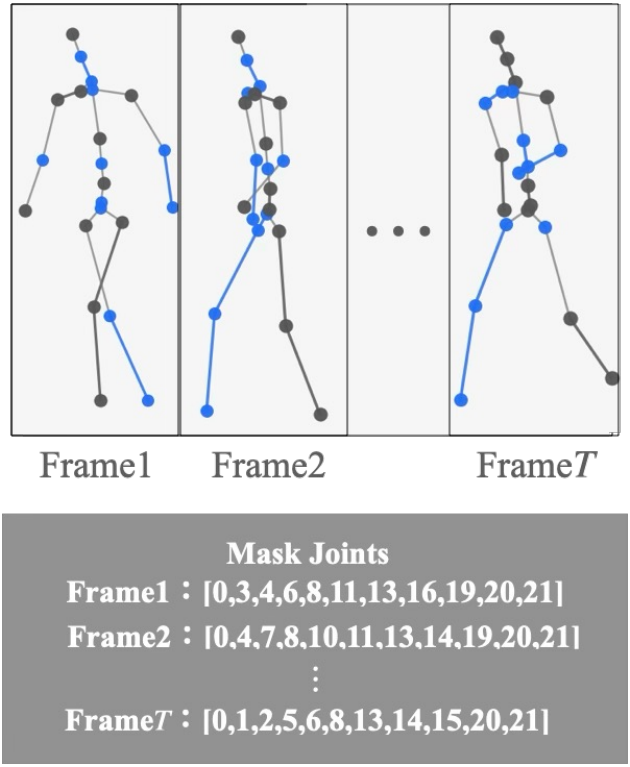}
    \caption{Random mask
    }
    \label{fig:mask_pattern_b}
  \end{subfigure}\hfill
  \begin{subfigure}[t]{0.15\textwidth}
    \centering
    \includegraphics[width=\linewidth]{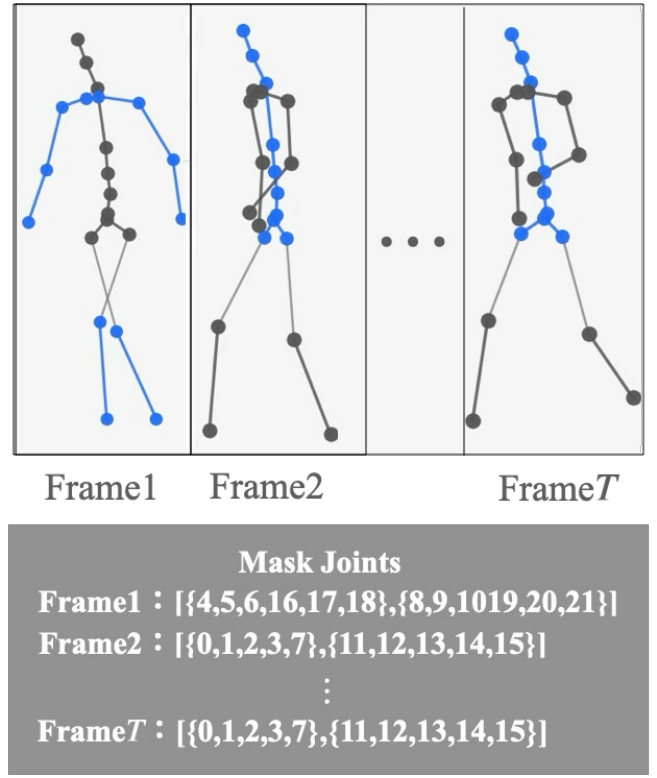}
    \caption{Body-Part mask}
    \label{fig:mask_pattern_c}
  \end{subfigure}
\caption{Illustration of three mask patterns used for self-supervised skeleton pretraining:
  (a) \textbf{Temporally Consistent} masks the same joints across all frames;
  (b) \textbf{Random} masks joints independently at each frame;
  (c) \textbf{Body-Part} masks multiple joints in the same body part together.}
  \label{fig:mask_pattern}
\end{figure}

\subsubsection{Masking Strategy}
In real-world scenarios, missing skeleton observations vary in both
temporal persistence and spatial extent.
To reflect these differences,
we design three masking strategies for the input skeleton sequence
$\mathbf{S} \in \mathbb{R}^{T \times N \times C}$,
as illustrated in Fig.~\ref{fig:mask_pattern}.
These strategies cover a range of missing patterns,
from short-term local dropouts to long-term large-scale occlusions.
For all masking strategies, masked joint-time locations are replaced with zeros.
This zero-filled representation serves as an explicit indicator of missing inputs under partial observations.

\noindent\textbf{Temporal Consistent Mask.}
This strategy applies an identical mask pattern across all frames
to a randomly selected subset of joints
(Fig.~\ref{fig:mask_pattern_a}).
The mask simulates long-term occlusions or tracking failures, with missing observations persisting over time.

\noindent\textbf{Random Mask.}
In this strategy, joints to be masked are independently sampled at each frame
(Fig.~\ref{fig:mask_pattern_b}).
This setting models sporadic detection failures and short-term occlusions,
exposing the encoder to temporally inconsistent missing patterns.

\noindent\textbf{Body-Part Mask.}
Following SkeletonMAE~\cite{10376785},
this strategy masks multiple joints simultaneously based on predefined body parts (Fig.~\ref{fig:mask_pattern_c}).
This mask simulates severe occlusions caused by obstacles or other agents,
with entire body regions becoming unobservable.
In our implementation, the skeleton is divided into four parts:
\textit{center\_upper}, \textit{center\_lower}, \textit{right\_limb}, and \textit{left\_limb}.

\subsubsection{Reconstruction Loss}
The objective of the self-supervised learning framework is to directly reconstruct masked joint coordinates.
To measure the discrepancy between the reconstructed skeleton sequence and the original input,
we employ the mean squared error (MSE) as the reconstruction loss.
Let the input skeleton sequence be
$\mathbf{S} \in \mathbb{R}^{T \times N \times C}$
and the reconstructed sequence be
$\mathbf{Y} \in \mathbb{R}^{T \times N \times C}$.
The reconstruction loss is defined as:
\begin{equation}
\mathcal{L}_{\mathrm{rec}}
=
\frac{1}{TN}
\left\|
\mathbf{S}
-
\mathbf{Y}
\right\|_2^2 .
\end{equation}

\subsection{Integration into Human Trajectory Prediction}
We integrate the skeleton encoder pretrained in a self-supervised manner
into a trajectory prediction model.

\subsubsection{Integration Strategy}
\begin{figure}[t]
    \centering
    \includegraphics[width=0.5\textwidth]{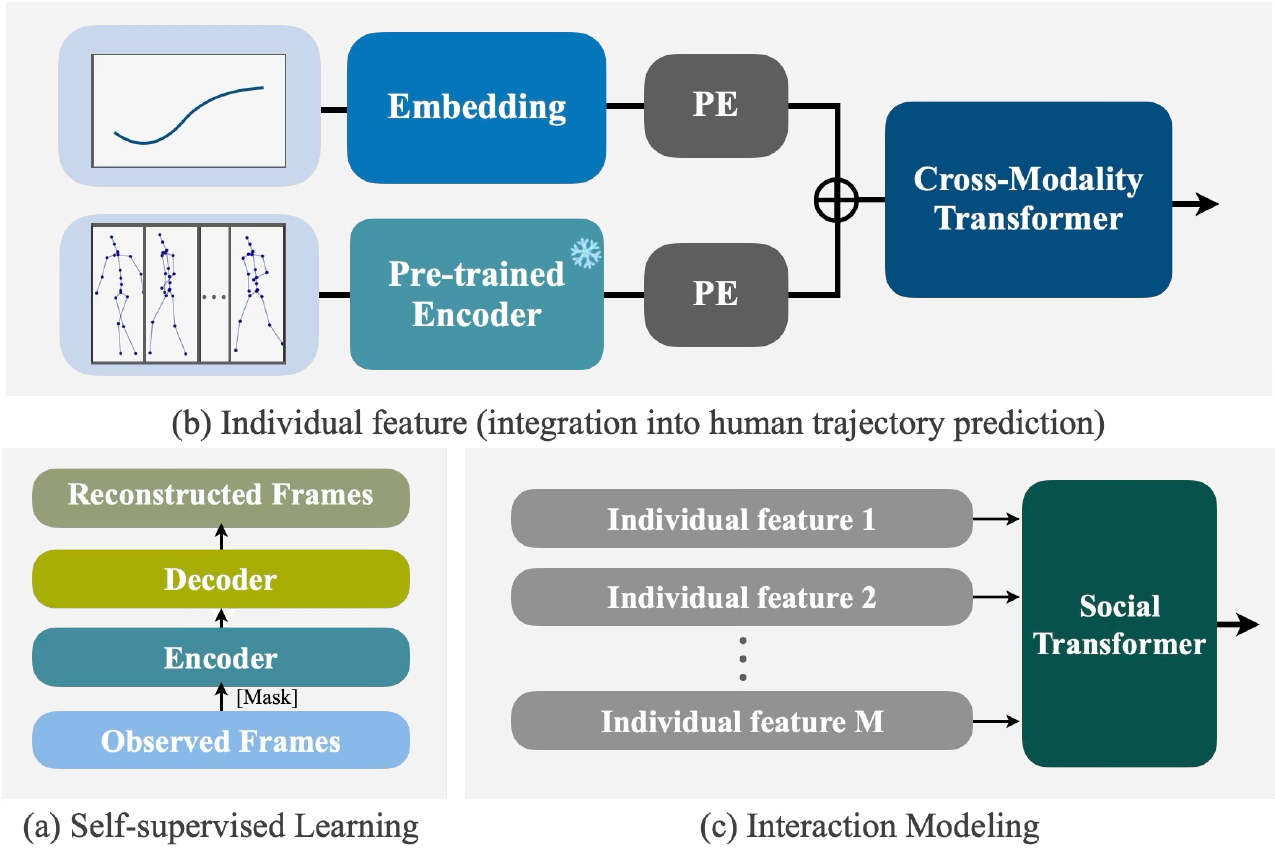}
\caption{
Overall architecture of the proposed framework.
(a) \textbf{Self-supervised pretraining}: a skeleton encoder is pretrained by reconstructing masked joints from partially observed skeleton sequences.
(b) \textbf{Individual feature extraction}: observed trajectories are embedded, and skeleton sequences are encoded by the pretrained (frozen) encoder. Positional encoding (PE) is added to each stream, the resulting tokens are concatenated, and a Cross-Modality Transformer produces an agent-wise representation.
(c) \textbf{Interaction modeling}: the agent-wise representations are processed by a Social Transformer to model inter-agent interactions.
}
    \label{fig:methods}
\end{figure}

As illustrated in Fig.~\ref{fig:methods},
we build our trajectory prediction model on the Social-TransMotion~\cite{saadatnejad2024socialtransmotion} framework,
which maps multiple modalities into a shared feature space via a two-stage architecture.
A \emph{Cross-Modality Transformer} integrates modality-specific motion features,
and a \emph{Social Transformer} models interactions among agents.
In the original Social-TransMotion framework,
each modality is embedded by a linear projection with positional encoding.
In contrast, we keep the trajectory embedding unchanged and replace only the 3D skeleton embedding
with a skeleton encoder $G_E$ pretrained via self-supervised learning.
During downstream trajectory-prediction training, the pretrained skeleton encoder $G_E$ is kept frozen unless otherwise stated.
That is, only the trajectory prediction modules are optimized in the downstream stage.

Let the input 3D skeleton sequence be $\mathbf{S} \in \mathbb{R}^{T \times N \times C}$,
where $T$ is the number of observed frames, $N$ is the number of joints, and $C$ is the coordinate dimension.
We encode $\mathbf{S}$ into a high-level representation using the pretrained skeleton encoder $G_E$,
and then apply positional encoding:
\begin{align}
\mathbf{H}^{(\mathrm{pose})}
&= G_E(\mathbf{S}),
\qquad
\mathbf{H}^{(\mathrm{pose})} \in \mathbb{R}^{T \times N \times D}, \\
\mathbf{z}^{(\mathrm{pose})}_{t,n}
&=
\mathrm{PE}\!\left(
\mathbf{h}^{(\mathrm{pose})}_{t,n}
\right),
\qquad
\mathbf{z}^{(\mathrm{pose})}_{t,n} \in \mathbb{R}^{D}.
\end{align}
Here, $\mathbf{h}^{(\mathrm{pose})}_{t,n} \in \mathbb{R}^{D}$ denotes the skeleton feature vector
for joint $n$ at time step $t$, and $\mathrm{PE}(\cdot)$ is the positional encoding based on temporal and joint indices.

While residual connections are sometimes used in module transfer to preserve the original embedding,
we intentionally omit such a skip path so that skeleton tokens are derived solely from the pretrained encoder.
Adding the original linear skeleton projection can allow the model to bypass the encoder
and directly rely on corrupted joint coordinates, undermining robustness under missingness.
Therefore, we fully replace the skeleton embedding with the encoder.

\subsubsection{Overall Architecture}

Our trajectory prediction model follows the overall design of Social-TransMotion~\cite{saadatnejad2024socialtransmotion}.
In this framework, the Cross-Modality Transformer integrates motion representations from multiple modalities,
while the Social Transformer models social interactions among agents.

In our setting, the modality set is restricted to trajectory (traj)
and 3D skeleton (pose) inputs.
The embedding of the 3D skeleton modality is replaced by the pretrained skeleton encoder,
while the rest of the framework remains unchanged.
Let $T_{\mathrm{obs}}$ and $T_{\mathrm{pred}}$ denote the lengths of the observation and prediction horizons, respectively.

For agent $i$, the embedded trajectory and skeleton sequences over the observation window are defined as
\begin{align}
\mathbf{Z}^{c}_i
&=
\bigl(
\mathbf{z}^{(c)}_{i,1}, \ldots,
\mathbf{z}^{(c)}_{i,T_{\mathrm{obs}}}
\bigr)
\in \mathbb{R}^{T_{\mathrm{obs}} \times e^{c} \times D},
\end{align}
where $e^{c}$ represents the number of elements in modality $c$.

Following Social-TransMotion,
the Cross-Modality Transformer takes as input a latent query sequence
$\mathbf{Q}_i \in \mathbb{R}^{T_{\mathrm{pred}} \times D}$
corresponding to the prediction horizon,
together with the modality-specific embeddings.
The output of the Cross-Modality Transformer is given by
\begin{equation}
\mathbf{mQ}_i,\; \mathbf{mZ}^{c}_i
=
\mathrm{CMT}\!\left(
\mathbf{Q}_i,\;
\mathbf{Z}^{c}_i,\;
c \in \{\mathrm{traj}, \mathrm{pose}\}
\right).
\end{equation}
Here, $\mathbf{mQ}_i \in \mathbb{R}^{T_{\mathrm{pred}} \times D}$ denotes the cross-modal query representation,
obtained by attending to the modality-specific embeddings within the Cross-Modality Transformer,
thereby aggregating information from both trajectory and skeleton cues.

Following Social-TransMotion,
the cross-modal representation fed into the Social Transformer is defined as
\begin{equation}
\mathbf{mZ}^{\mathrm{M}}_i
=
\mathrm{Concat}\!\left(
\mathbf{mQ}_i,\;
\mathbf{mZ}^{\mathrm{traj}}_i
\right)
\in \mathbb{R}^{(T_{\mathrm{pred}}+T_{\mathrm{obs}}) \times D}.
\end{equation}
This design incorporates skeleton information into trajectory representations
without introducing explicit skeleton tokens,
thus preserving the original input structure of the Social Transformer.

The Social Transformer output is denoted as
$\mathbf{SM}_i=\mathrm{Concat}\!\left(\mathbf{SM}^{\mathrm{T}}_i,\mathbf{SM}^{\mathrm{Q}}_i\right)$,
where $\mathbf{SM}^{\mathrm{T}}_i$ and $\mathbf{SM}^{\mathrm{Q}}_i$ correspond to
$\mathbf{mZ}^{\mathrm{traj}}_i$ and $\mathbf{mQ}_i$, respectively.
Finally, the future trajectory of the target agent ($i=1$) is predicted from the query-stream output of the Social Transformer:
\begin{equation}
\hat{\mathbf{Y}}_1 = \mathrm{MLP}\!\left(\mathbf{SM}^{\mathrm{Q}}_1\right)
\in \mathbb{R}^{T_{\mathrm{pred}} \times 2}.
\end{equation}

\subsubsection{Loss Function}
We use the mean squared error (MSE) between the predicted and ground-truth trajectories as the training loss:
\begin{equation}
\mathcal{L}_{\mathrm{traj}}
=
\frac{1}{T_{\mathrm{pred}}}
\sum_{t=1}^{T_{\mathrm{pred}}}
\left\|
\hat{\mathbf{y}}_{t}
-
\mathbf{y}_{t}
\right\|_2^2 .
\end{equation}
Here,
$\mathbf{y}_{t}, \hat{\mathbf{y}}_{t} \in \mathbb{R}^{2}$
denote the ground-truth and predicted pedestrian positions at time step $t$, respectively.
 \section{Experiments}
We conduct experiments in two stages: (i) self-supervised skeleton representation learning and
(ii) fine-tuning for human trajectory prediction.
In the first stage, we evaluate reconstruction quality and representation consistency under joint masking.
In the second stage, we assess trajectory prediction performance in partial-observation scenarios.

\subsection{Datasets}
We conduct experiments on the JTA dataset~\cite{fabbri2018learning}, a large-scale synthetic dataset of pedestrian behaviors in complex crowd scenes.
JTA simulates realistic urban environments and contains frequent occlusions.
Following the protocol of SocialTransMotion~\cite{saadatnejad2024socialtransmotion}, we use $T_{\mathrm{obs}}=9$ observed frames and predict $T_{\mathrm{pred}}=12$ future frames at 2.5\,fps (i.e., 3.6\,s observation and 4.8\,s prediction).
For self-supervised pretraining, we also use 9-frame skeleton sequences for consistency with the downstream observation length.

\subsection{Self-Supervised Skeleton Representation Learning}
\label{subsec:self-supervised}

\subsubsection{Implementation Details}
For self-supervised pretraining on skeleton sequences,
the encoder-decoder model is optimized using Adam
with an initial learning rate of $1.5 \times 10^{-4}$.
PReLU is used as the activation function.
The batch size is set to 1024, and the model is trained for 50 epochs.
All pretraining experiments are implemented in PyTorch
and conducted on a single NVIDIA RTX 6000 GPU.

\subsubsection{Evaluation Metrics}
To evaluate the representations learned by self-supervised pretraining,
we use two complementary metrics:
(i) reconstruction accuracy in 3D joint space, and
(ii) stability of encoder features when joints are masked.

\paragraph{Joint Position Error (MPJPE)}
We measure reconstruction accuracy using the mean per-joint position error (MPJPE) in meters (m),
defined as the average $\ell_2$ distance between the reconstructed joint positions
$\mathbf{y}_{t,i}$ and the ground-truth positions $\mathbf{x}_{t,i}$ over all joints and time steps:
\begin{equation}
\mathrm{MPJPE}
=
\frac{1}{T |V|}
\sum_{t=1}^{T}
\sum_{v_i \in V}
\left\|
\mathbf{x}_{t,i}
-
\mathbf{y}_{t,i}
\right\|_2,
\end{equation}
where $t$ is the time index and $V$ is the set of joints.

\paragraph{Feature Consistency under Joint Masking}
To quantify how consistent the encoder features are under joint masking,
we compute cosine similarity between latent representations extracted from
the unmasked input, $\mathbf{h}_{t,i}$, and the masked input, $\tilde{\mathbf{h}}_{t,i}$.
\begin{equation}
\mathrm{CosSim}
=
\frac{1}{T |V|}
\sum_{t=1}^{T}
\sum_{v_i \in V}
\frac{
\mathbf{h}_{t,i}^\top \tilde{\mathbf{h}}_{t,i}
}{
\|\mathbf{h}_{t,i}\|_2
\,
\|\tilde{\mathbf{h}}_{t,i}\|_2
}.
\end{equation}
Higher cosine similarity indicates more stable representations under joint masking.

\begin{table}[t]
\centering
\caption{
Cross-evaluation of three masking strategies (Fig.~\ref{fig:mask_pattern}).
Each model is trained with one strategy and evaluated under all three test-time masks
(Temporally-Consistent, Random, Body-Part) with a fixed test mask ratio of 0.5.
Bold indicates the best result within each training strategy (min MPJPE / max cosine similarity).
}
\label{tab:mask_strategy_cross}
\resizebox{\linewidth}{!}{
\begin{tabular}{l|l|cc}
\hline
\textbf{Train} & \textbf{Test} &
\textbf{MPJPE} &
\textbf{Cos Sim}\\
\hline
\multirow{3}{*}{Temporally-Consistent}
& Temporally-Consistent  & \textbf{0.058} & \textbf{0.841} \\
& Random                 & 0.201 & 0.833 \\
& Body-Part              & 0.215 & 0.821 \\
\hline
\multirow{3}{*}{Random}
& Temporally-Consistent  & 0.062 & 0.912 \\
& Random                 & \textbf{0.036} & \textbf{0.935} \\
& Body-Part              & 0.047 & 0.928 \\
\hline
\multirow{3}{*}{Body-Part}
& Temporally-Consistent  & 0.213 & 0.805 \\
& Random                 & 0.183 & 0.846 \\
& Body-Part              & \textbf{0.051} & \textbf{0.882} \\
\hline
\end{tabular}}
\end{table}

\subsubsection{Effect of Masking Design}
\label{sec:ablation_ssl}

We examine how reconstruction-based pretraining behaves under different masking designs.
Unless otherwise noted, we use Random masking with a training mask ratio $r_{\mathrm{train}}=0.5$ and a 3-layer ST-GCN encoder.
In this section, we vary one factor at a time---(i) masking strategy and (ii) training mask ratio---while keeping all other settings fixed.
We report reconstruction accuracy (MPJPE) and representation consistency measured by cosine similarity between latent features from clean and masked inputs.

\paragraph{Effect of Masking Strategy}

We analyze how the masking strategy affects reconstruction accuracy and how well a model transfers to different test-time masking patterns.
We consider three strategies (Fig.~\ref{fig:mask_pattern}): Temporally-Consistent, Random, and Body-Part.

Table~\ref{tab:mask_strategy_cross} reports cross-evaluation results, where each model is trained with one strategy and evaluated under all three test-time masks with $r_{\mathrm{test}}=0.5$.
Across the three strategies, the lowest MPJPE is obtained when the training and test masks match, while mismatched combinations lead to larger errors.
For instance, the Temporally-Consistent model increases its MPJPE from 0.058 (matched) to 0.201 and 0.215 under Random and Body-Part tests, respectively.
A similar pattern is observed for the Body-Part model (0.051 matched vs.\ 0.213/0.183 under Temporally-Consistent/Random tests).
This suggests that structured masking can bias the model toward specific temporal or spatial missing structures, which may not carry over to other masking patterns.
In contrast, Random masking yields low MPJPE (0.036–0.062) together with high cosine similarity (0.912–0.935) across test-time masks, indicating that the encoder captures informative motion features that remain consistent even when joints are missing.
Based on these observations, we adopt Random masking as the default strategy for downstream trajectory prediction.

\begin{figure}[t]
  \centering
  \begin{subfigure}[t]{0.51\linewidth}
    \centering
    \includegraphics[width=\linewidth]{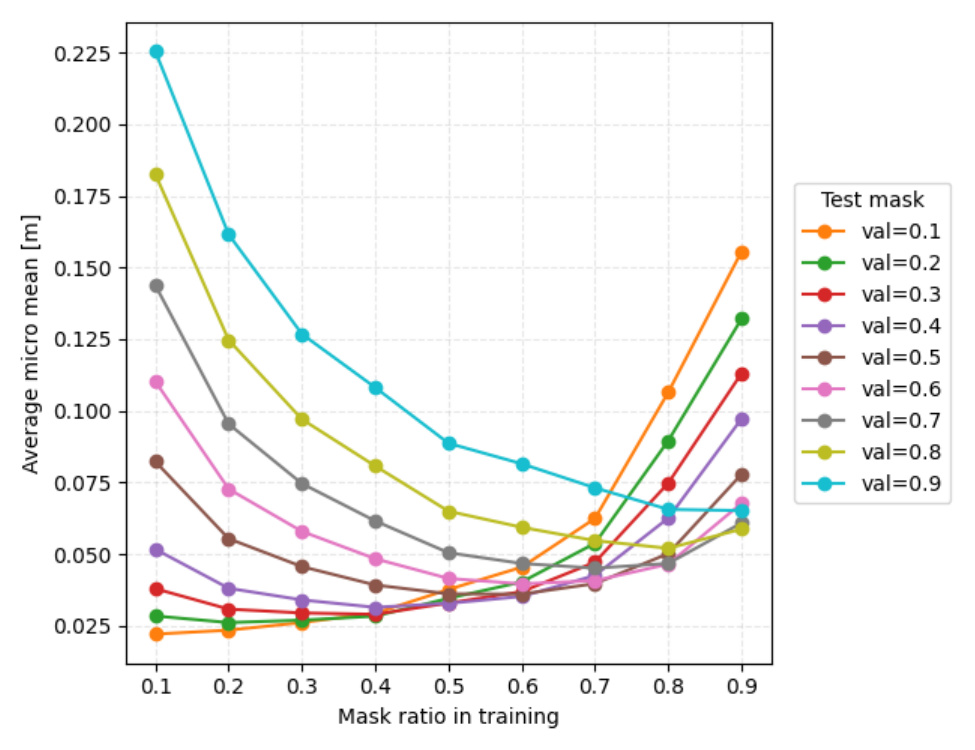}
    \caption{Line plot.}
    \label{fig:mask_rate-line}
  \end{subfigure}\hfill
  \begin{subfigure}[t]{0.47\linewidth}
    \centering
    \includegraphics[width=\linewidth]{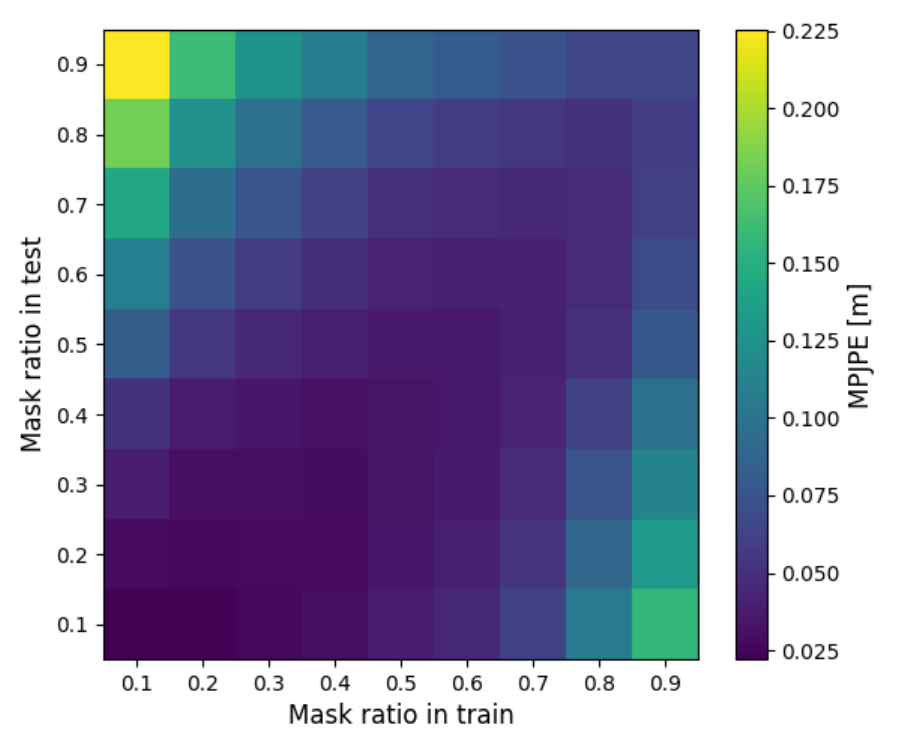}
    \caption{Heatmap.}
    \label{fig:mask_rate-heatmap}
  \end{subfigure}
  \caption{
    Reconstruction MPJPE under varying training and test-time mask ratios.
    (a) MPJPE versus the training mask ratio $r_{\mathrm{train}}$, with one curve for each test-time mask ratio $r_{\mathrm{test}}$.
    (b) Heatmap over $(r_{\mathrm{train}}, r_{\mathrm{test}})$, where darker colors indicate lower MPJPE.
  }
  \label{fig:mask_rate}
\end{figure}

\begin{table}[t]
\centering
\caption{
Reconstruction performance under different training mask ratios.
Each model is evaluated across test-time mask ratios $r_{\mathrm{test}} \in \{0.1,0.2,\dots,0.9\}$.
We report the mask-rate-averaged MPJPE (mean over $r_{\mathrm{test}}$).
}
\label{tab:mask_rate_stats}
\resizebox{\linewidth}{!}{
\begin{tabular}{c|ccccccc}
\hline
\textbf{Mask ratio in train} 
& \textbf{0.2}& \textbf{0.3} & \textbf{0.4} & \textbf{0.5} & \textbf{0.6} & \textbf{0.7}&\textbf{0.8} \\
\hline
Mask-rate-avg
&0.070 & 0.058 & 0.051 & \textbf{0.046} & 0.047 & 0.051& 0.066 \\
\hline
\end{tabular}
}
\end{table}

\paragraph{Effect of Masking Ratio}

Fig.~\ref{fig:mask_rate} shows reconstruction MPJPE under varying masking ratios.
In Fig.~\ref{fig:mask_rate-line}, the horizontal axis denotes the training mask ratio $r_{\mathrm{train}}$, and each curve corresponds to a test-time mask ratio $r_{\mathrm{test}} \in \{0.1,0.2,\dots,0.9\}$.
Fig.~\ref{fig:mask_rate-heatmap} presents the same results as a heatmap for compact visualization.

Overall, using overly small or overly large training mask ratios increases sensitivity to the test-time masking level.
When $r_{\mathrm{train}}$ is small, most joints remain visible during pretraining, making the reconstruction task relatively easy.
As a result, the encoder receives limited supervision for inferring missing joints, and the error increases markedly under severe test-time masking.
Conversely, when $r_{\mathrm{train}}$ is large, the amount of observed evidence becomes insufficient, which hinders recovery of fine-grained joint configurations and leads to over-smoothed reconstructions.
This causes elevated errors even under low test-time masking.

Table~\ref{tab:mask_rate_stats} summarizes these results using the mask-rate-averaged MPJPE, computed by averaging MPJPE over the nine $r_{\mathrm{test}}$ values for each $r_{\mathrm{train}}$.
The averaged error exhibits a U-shaped trend, reaching its minimum at $r_{\mathrm{train}}=0.5$.
These results suggest that a non-extreme training mask ratio provides a good balance between preserving sufficient visible joints and learning to infer missing ones; therefore, we use $r_{\mathrm{train}}=0.5$ as the default setting.

\begin{figure}[t]
  \centering
  \includegraphics[width=1.0\linewidth]{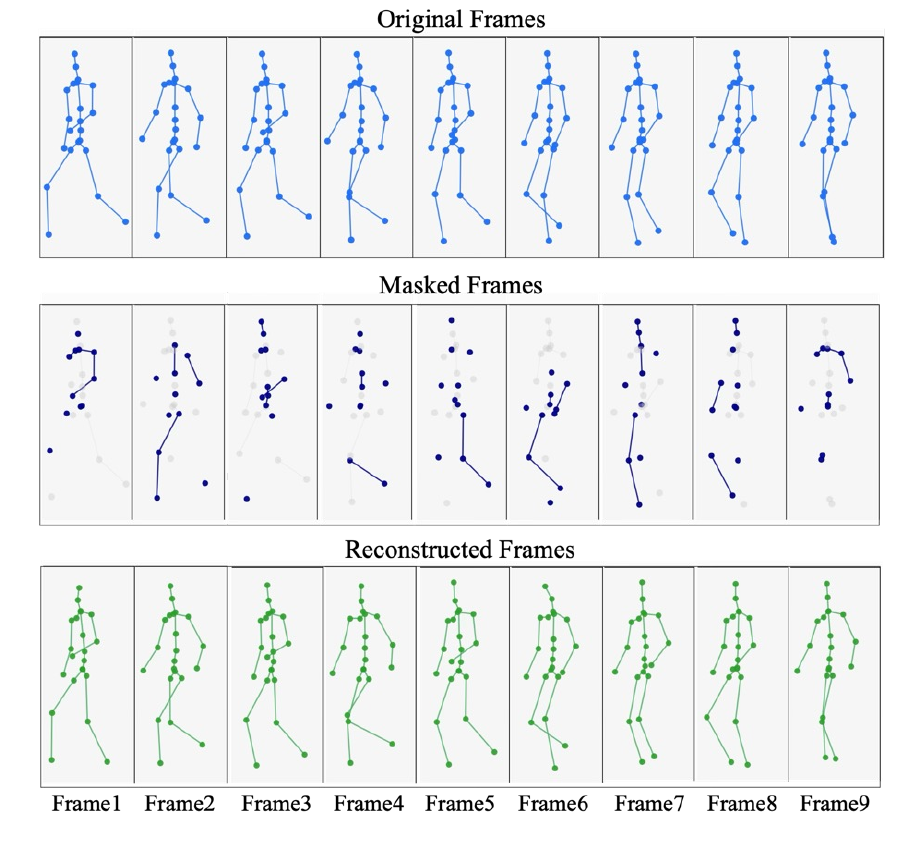}
  \caption{
Reconstruction example at a representative test-time mask ratio ($r_{\mathrm{test}}=0.5$).
The model reconstructs a coherent body structure and preserves the overall motion trend across frames.
  }
  \label{fig:reconstruct_example}
\end{figure}

\subsubsection{Visualization}
\label{sec:qualitative_reconstruction}

Fig.~\ref{fig:reconstruct_example} shows a representative reconstruction example at $r_{\mathrm{test}}=0.5$ using the default pretraining setting (Random masking with $r_{\mathrm{train}}=0.5$).
Despite missing observations, the reconstruction closely follows the ground-truth motion, with only minor joint-level deviations.
The model also preserves a coherent body configuration and maintains the overall motion trend across frames, rather than collapsing to a time-invariant or overly averaged pose.
Overall, this qualitative example matches the quantitative trends in Fig.~\ref{fig:mask_rate}.

\subsection{Fine-tuning for Human Trajectory Prediction}

\subsubsection{Implementation Details}
For downstream human trajectory prediction,
we follow the configuration of Social-TransMotion.
The Cross-Modality Transformer consists of 6 layers with 4 attention heads,
while the Social Transformer is composed of 3 layers with 4 attention heads.
The model dimension is fixed to 128 for all components.
Adam is used as the optimizer with an initial learning rate of $1 \times 10^{-4}$.
The learning rate is decayed by a factor of 0.1
after 80\% of the total 50 training epochs.

\subsubsection{Evaluation Metrics}
We evaluate trajectory prediction performance using Average Displacement Error (ADE)
and Final Displacement Error (FDE), two standard metrics.

\paragraph{Average Displacement Error (ADE)}
ADE measures the mean Euclidean distance between the predicted positions
$\hat{\mathbf{y}}_t$ and the ground-truth positions $\mathbf{y}_t$
throughout prediction.
\begin{equation}
\mathrm{ADE}
=
\frac{1}{T_{\mathrm{pred}}}
\sum_{t=1}^{T_{\mathrm{pred}}}
\left\|
\mathbf{y}_{t}
-
\hat{\mathbf{y}}_{t}
\right\|_2 .
\end{equation}

\paragraph{Final Displacement Error (FDE)}
FDE measures the Euclidean distance between
the predicted and ground-truth positions
at the final prediction timestep.

\begin{equation}
\mathrm{FDE}
=
\left\|
\mathbf{y}_{T_{\mathrm{pred}}}
-
\hat{\mathbf{y}}_{T_{\mathrm{pred}}}
\right\|_2 .
\end{equation}

\begin{table*}[t]
\centering
\caption{
Comparison under random joint masking.
Missingness levels are grouped as Clean ($0.0$), Low ($0.2$), Moderate ($0.4$), and High ($0.6$), where the value in parentheses denotes the mask rate.
Each cell reports ADE / FDE, followed by the degradation rate (\%) relative to the clean condition of the same method.
Bold indicates the best ADE / FDE at each missingness level.
}
\label{tab:joint_masking_unified}
\resizebox{\textwidth}{!}{
\begin{tabular}{lcccc}
\hline
\textbf{Method}
& \textbf{Clean (0.0)} & \textbf{Low (0.2)} & \textbf{Moderate (0.4)} & \textbf{High (0.6)} \\
\hline
Baseline (Standard)
& 0.920 / 1.884 (+0.0 / +0.0\%)
& 0.940 / 1.916 (+2.2 / +1.7\%)
& 0.969 / 1.957 (+5.3 / +3.9\%)
& 1.050 / 2.090 (+14.0 / +10.9\%) \\

Baseline (Reconstruction)
& 0.920 / 1.884 (+0.0 / +0.0\%)
& 0.927 / 1.892 (+0.8 / +0.4\%)
& 0.940 / 1.907 (+2.1 / +1.2\%)
& 0.973 / \textbf{1.939} (+5.7 / +2.9\%) \\

Baseline (Corruption-trained)
& 0.956 / 1.948 (+0.0 / +0.0\%)
& 0.953 / 1.942 (-0.3 / -0.3\%)
& 0.955 / 1.946 (-0.1 / -0.1\%)
& \textbf{0.960} / 1.955 (+0.4 / +0.4\%) \\
\hline
Ours
& \textbf{0.898 / 1.832} (+0.0 / +0.0\%)
& \textbf{0.913 / 1.853} (+1.7 / +1.1\%)
& \textbf{0.934 / 1.884} (+4.0 / +2.8\%)
& 1.017 / 2.058 (+13.3 / +12.3\%) \\
\hline
\end{tabular}
}
\end{table*}

\begin{table}[t]
\centering
\caption{
Analysis of skeleton contribution.
We quantify the contribution of skeleton cues by zeroing the 3D pose part in the Cross-Modality Transformer input and measuring the resulting performance drop.
A larger drop ($\Delta$) indicates stronger reliance on skeleton information.
}
\label{tab:skeleton_contribution_analysis}
\resizebox{\linewidth}{!}{%
\begin{tabular}{lccc}
\hline
\textbf{Method} & \textbf{Clean} & \textbf{Skeleton disabled} & $\mathbf{\Delta}$ \textbf{(ADE / FDE)} \\
\hline
Baseline (Standard / Reconstruction) & 0.920 / 1.884 & 1.563 / 3.186 & +0.643 / +1.301 \\
Baseline (Corruption-trained) & 0.956 / 1.948 & 1.239 / 2.355 & +0.283 / +0.407 \\
Ours & 0.898 / 1.832 & 2.389 / 4.789 & +1.491 / +2.957 \\
\hline
\end{tabular}%
}
\end{table}

\subsubsection{Robustness Evaluation under Random Joint Masking}
\label{sec:robustness_eval}

We evaluate robustness to test-time joint missingness by varying the masking ratio.
We compare our method with the following three baselines:
\begin{itemize}
    \item \textbf{Baseline (Standard)}: the original Social-TransMotion.
    \item \textbf{Baseline (Reconstruction)}: a reconstruction-preprocessing baseline that uses the reconstruction model trained in Section~\ref{subsec:self-supervised} to complete missing joints before prediction. The trajectory predictor uses the same weights as the Baseline (Standard).
    \item \textbf{Baseline (Corruption-trained)}: the same predictor architecture trained directly on corrupted skeleton inputs with a fixed random joint masking rate of $0.5$.
\end{itemize}

\paragraph{Results}
Table~\ref{tab:joint_masking_unified} summarizes the results under random joint masking.
\emph{Ours} achieves the best ADE/FDE from Clean to Moderate missingness ($0.0$--$0.4$), while \emph{Baseline (Reconstruction)} performs best at High missingness ($0.6$).
\emph{Baseline (Corruption-trained)} shows a nearly flat error profile across mask ratios, but its clean-input accuracy is lower than both \emph{Baseline (Standard)} and \emph{Ours}.

\paragraph{Discussion}
The two baselines represent complementary strategies.
\emph{Baseline (Corruption-trained)} trains the downstream predictor on corrupted skeleton inputs, which flattens the degradation trend but lowers clean-condition accuracy.
In contrast, \emph{Baseline (Reconstruction)} reconstructs missing joints before prediction,
which can reduce missingness effects.
However, clean-condition accuracy remains bounded by \emph{Baseline (Standard)} because the downstream predictor is identical in the two baselines.
Accordingly, both baselines outperform \emph{Ours} at High missingness and exhibit smaller degradation rates at low-to-moderate missingness.
In the practically important low-to-moderate missingness regime, \emph{Ours} still achieves the best ADE/FDE.

\subsubsection{Skeleton Reliance Diagnostics}
Robustness cannot be compared only from ranking based on degradation rates (Table~\ref{tab:joint_masking_unified}), 
because the three compared methods can rely on skeleton cues to different degrees.
To quantify skeleton reliance, we disable the skeleton (3D pose) part of the Cross-Modality Transformer input on clean data and measure the resulting performance drop.
A larger drop indicates stronger reliance on skeleton information.

\paragraph{Results}
Table~\ref{tab:skeleton_contribution_analysis} reports the performance drop when skeleton cues are disabled.
\emph{Ours} exhibits a larger drop than \emph{Baseline (Standard)} ($+1.491/+2.957$ vs.\ $+0.643/+1.301$ in ADE/FDE), indicating stronger reliance on skeleton cues.
In contrast, \emph{Baseline (Corruption-trained)} shows a smaller drop, suggesting weaker effective reliance.
This gap implies that the two baselines can appear more robust when robustness is judged by degradation rate, especially at high missingness.
Since \emph{Ours} relies more on skeleton cues, \emph{Ours} benefits most when these cues are sufficiently available, i.e., from Clean to Moderate missingness.

To compare robustness as fairly as possible, differences in skeleton reliance should be minimized.
Appendix~\ref{sec:appendix_skeleton_contribution} provides controlled comparisons with matched skeleton dependence.

\paragraph{Discussion}
Stronger skeleton reliance can improve clean-input accuracy, but it can also increase sensitivity when joints are missing.
This trade-off makes it difficult to improve clean accuracy and missingness robustness 
simultaneously in skeleton-aided trajectory prediction.
Despite stronger skeleton reliance, \emph{Ours} remains competitive in degradation at low-to-moderate missingness and achieves the best ADE/FDE from Clean to Moderate missingness.
This result suggests that the proposed self-supervised pretraining improves robustness by strengthening skeleton motion representations under partial observations, rather than by reducing skeleton reliance.


\subsection{Visualization}
\label{sec:visualization}

\begin{figure}[t]
  \centering
  \begin{subfigure}[t]{0.49\linewidth}
    \centering
    \includegraphics[width=\linewidth]{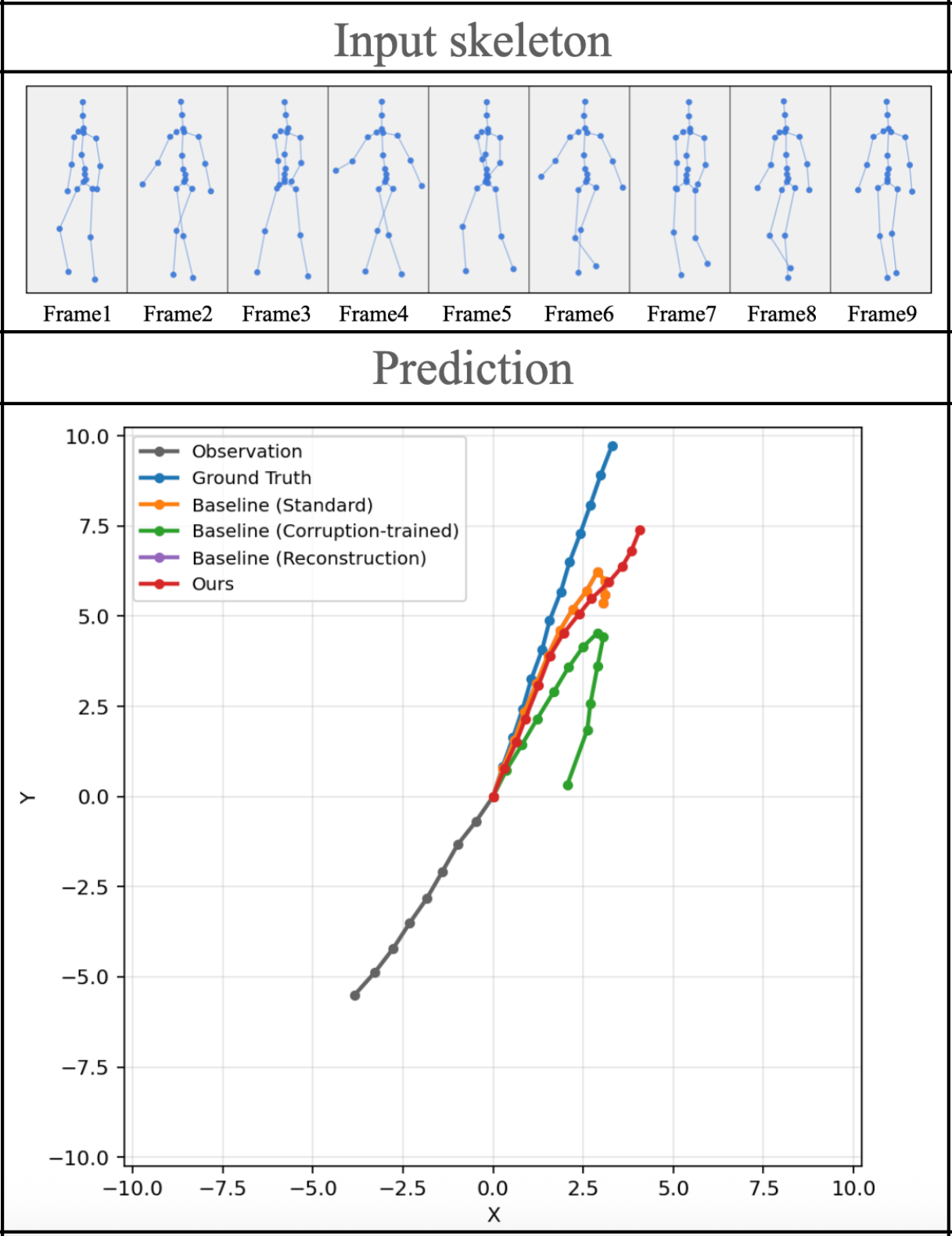}
    \caption{Clean (0.0)}
    \label{fig:121}
  \end{subfigure}\hfill
  \begin{subfigure}[t]{0.49\linewidth}
    \centering
    \includegraphics[width=\linewidth]{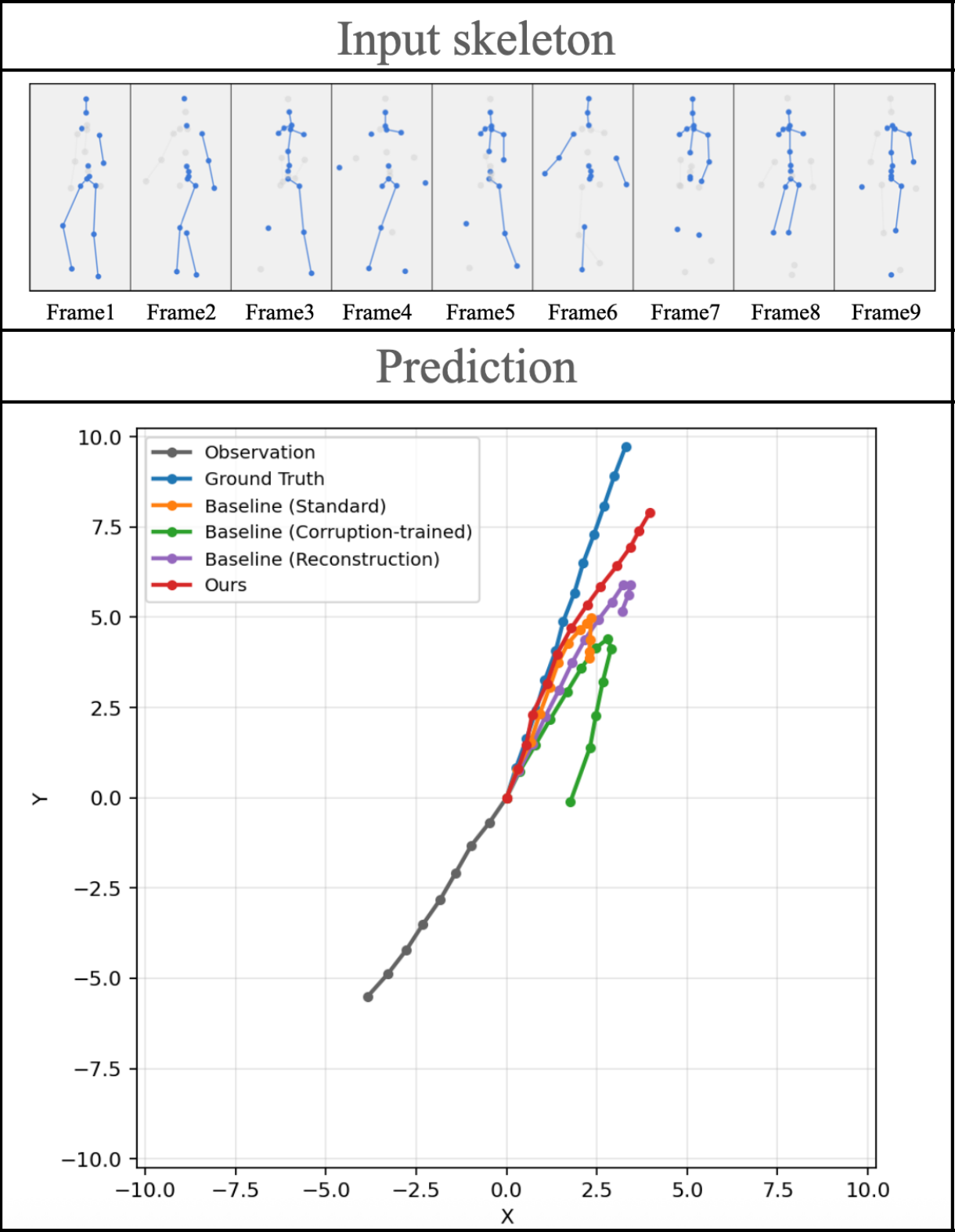}
    \caption{Moderate missingness (0.4)}
    \label{fig:qualitative_straight_mask}
  \end{subfigure}
\caption{
Straight-trajectory case under clean and masked inputs.
This example is consistent with the quantitative trends: \emph{Ours} remains stable, while the baselines show either masking-induced degradation or lower clean accuracy.
}
  \label{fig:qualitative_straight}
\end{figure}

\begin{figure}[t]
  \centering
  \begin{subfigure}[t]{0.49\linewidth}
    \centering
    \includegraphics[width=\linewidth]{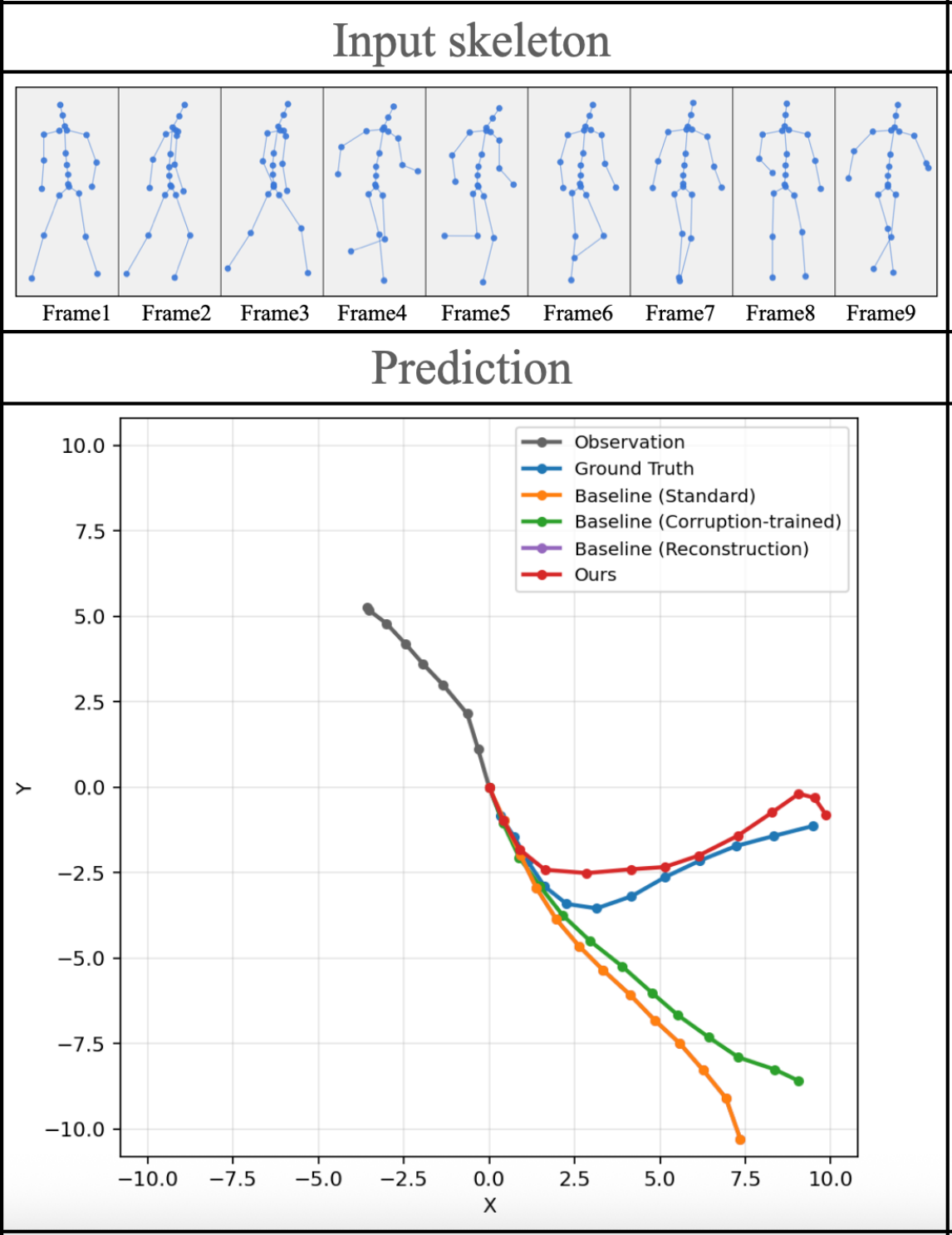}
    \caption{Clean (0.0)}
    \label{fig:qualitative_curve_clean}
  \end{subfigure}\hfill
  \begin{subfigure}[t]{0.49\linewidth}
    \centering
    \includegraphics[width=\linewidth]{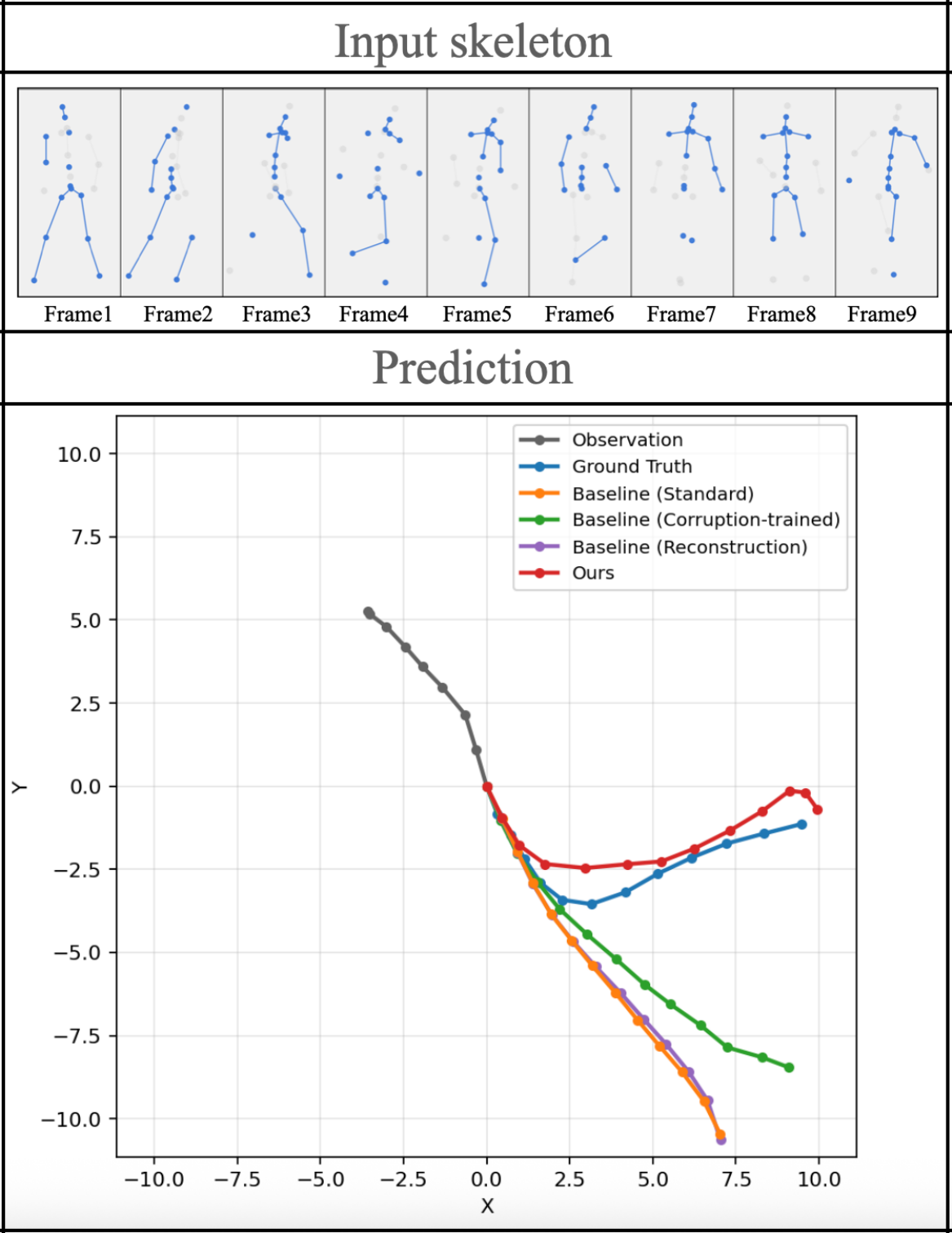}
    \caption{Moderate missingness (0.4)}
    \label{fig:qualitative_curve_mask}
  \end{subfigure}
\caption{
Curved-trajectory case under clean and masked inputs.
This is a skeleton-sensitive case: the baselines often fail to capture path curvature even under clean input, whereas \emph{Ours} preserves the turning tendency under both clean and masked inputs.
}
  \label{fig:qualitative_curve}
\end{figure}

Figs.~\ref{fig:qualitative_straight} and~\ref{fig:qualitative_curve} provide qualitative comparisons on two representative scenes to illustrate the accuracy--robustness trade-off and how the proposed method mitigates it.
We compare four methods: \emph{Baseline (Standard)}, \emph{Baseline (Reconstruction)}, \emph{Baseline (Corruption-trained)}, and \emph{Ours}.
For each scene, we show predictions under clean input (0.0) and moderate missingness (0.4) with random joint masking.

\paragraph{Case 1: Straight trajectory}
The first example is a simple straight-motion scene and is consistent with the quantitative trends in Table~\ref{tab:joint_masking_unified}.
Under clean input, \emph{Baseline (Corruption-trained)} already shows lower accuracy, indicating the clean-accuracy cost of training directly on corrupted skeletons.
\emph{Baseline (Standard)} predicts well under clean input but degrades under masking, incorrectly predicting a turn after missing a pose cue.
\emph{Baseline (Reconstruction)} partially mitigates this degradation, but its performance remains bounded by the inherited downstream predictor.
In contrast, \emph{Ours} remains stable in both clean and masked conditions, suggesting that the proposed representation preserves motion-relevant skeleton cues while improving robustness to missing joints.

\paragraph{Case 2: Curved trajectory}
The second example is a curved-motion scene, where accurate prediction requires stronger use of skeletal cues such as body orientation and motion tendency.
In this more challenging case, the baselines often fail to capture path curvature even under clean input, producing overly straight forecasts.
By contrast, \emph{Ours} follows the curved trajectory under clean input and maintains this behavior under moderate missingness.
This supports that the proposed method not only uses skeleton cues more effectively, but also preserves their usefulness under partial observations.

Overall, these examples suggest that the proposed method improves robustness not by weakening skeleton reliance, but by learning skeleton representations that remain informative under missing joints.
This qualitative evidence is consistent with the quantitative results in Table~\ref{tab:joint_masking_unified}.

 \section{conclusion}
In this study, we proposed a framework based on self-supervised skeleton representation learning for robust human trajectory prediction from skeleton sequences.
In skeleton-aided trajectory prediction, there is an inherent trade-off between prediction accuracy and robustness to missing joints.
To address this issue, the proposed method pretrains skeleton representations via reconstruction learning from incomplete skeleton inputs, and then integrates the pretrained encoder into a downstream trajectory predictor.
This design aims to improve robustness under missing joints while preserving the ability to effectively exploit skeleton cues.
Experimental results showed that the proposed method consistently improves prediction accuracy from clean to moderate missingness, while remaining competitive under severe missingness. 
Additional skeleton-reliance analysis and qualitative comparisons further showed that these gains do not come from weakening skeleton usage. 
Instead, the gains come from preserving the usefulness of skeleton features under partial observations.
Overall, these findings suggest that the proposed method is a promising approach for achieving both high accuracy and robustness to missing joints in skeleton-aided trajectory prediction.

\section*{Acknowledgement}
This work was supported by JSPS KAKENHI Grant Number JP23K24914.

\bibliographystyle{plain}
\bibliography{refs}
 \clearpage
\setcounter{page}{1}
\maketitlesupplementary

\appendix

\begin{table}[t]
\centering
\caption{
Effect of encoder depth $L$ in the ST-GCN-based reconstruction encoder.
We report mask-rate-averaged MPJPE (lower is better), averaged over
test-time mask ratios $r_{\mathrm{test}} \in \{0.1,0.2,\dots,0.9\}$.
}
\label{tab:encoder_depth_ablation_en}
\setlength{\tabcolsep}{6pt}
\renewcommand{\arraystretch}{1.1}
\resizebox{\linewidth}{!}{
\begin{tabular}{lccccc}
\hline
\textbf{Depth} & 1 & 2 & 3 & 4  \\
\hline
\textbf{Mask-rate-avg MPJPE} & 0.058 & 0.047 & \textbf{0.046} & 0.051  \\
\hline
\end{tabular}}
\end{table}

\begin{table}[t]
\centering
\caption{
Effect of reconstruction loss design in self-supervised pretraining (\(r_{\mathrm{train}}=0.5\), \(r_{\mathrm{val}}=0.5\)).
Masked-only loss slightly improves masked-joint reconstruction, but severely degrades unmasked-joint reconstruction and representation consistency.
}
\label{tab:loss_design_ablation}
\setlength{\tabcolsep}{5pt}
\renewcommand{\arraystretch}{1.12}
\resizebox{\linewidth}{!}{
\begin{tabular}{lcccc}
\hline
\textbf{Loss design} &
\textbf{Masked} &
\textbf{Unmasked} &
\textbf{Average} &
\textbf{Cos sim} \\
\hline
All-joint loss   & 0.046 & 0.026 & 0.036 & 0.935 \\
Masked-only loss & 0.043 & 0.218 & 0.131 & 0.899 \\
\hline
\end{tabular}}
\end{table}

\section{Ablation Studies on Self-Supervised Skeleton Learning}
\label{ablation}
\subsection{Effect of Encoder Depth}

We compared reconstruction performance across different encoder depths. As shown in Table~\ref{tab:encoder_depth_ablation_en}, the 3-layer encoder achieved the lowest reconstruction error. Although the differences among encoders with $L \ge 2$ were relatively small, performance improved from 1 to 3 layers. On the other hand, the 4-layer encoder showed a slightly higher reconstruction error than the 3-layer encoder. These results indicate that simply increasing the depth does not always improve performance. Therefore, we adopt the 3-layer encoder.

\subsection{Effect of Reconstruction Loss Design}
\label{sec:appendix_loss_design}
We analyze the effect of the reconstruction loss design in self-supervised pretraining.
While masked modeling often uses a loss only on masked joints, our downstream predictor relies directly on the encoder's latent representations (without skip connections), so representation stability for both masked and unmasked joints is important.

We compare two loss designs under the same setting (\(r_{\mathrm{train}}=0.5\), \(r_{\mathrm{val}}=0.5\)): All-joint loss (loss on all joints) and Masked-only loss (loss on masked joints only).
As shown in Table~\ref{tab:loss_design_ablation}, masked-only loss slightly improves masked-joint error, but severely degrades unmasked-joint error, resulting in worse average error and lower cosine similarity between masked and unmasked representations.

In contrast, all-joint loss maintains lower errors on both joint subsets and yields more consistent latent representations.
Because our method uses these latent representations for downstream prediction, we adopt the all-joint reconstruction loss in all experiments.

\begin{figure}[t]
  \centering
  \begin{subfigure}[t]{0.49\linewidth}
    \centering
    \includegraphics[width=\linewidth]{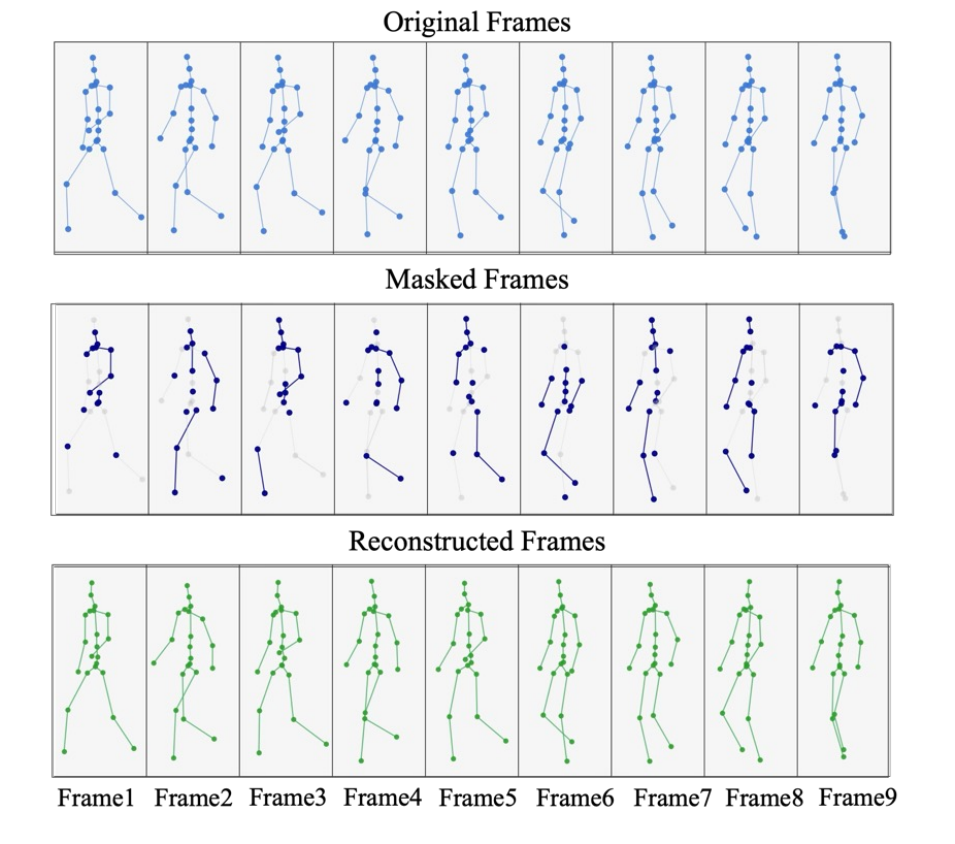}
    \caption{$r_{\mathrm{val}}=0.4$}
    \label{fig:recon_ex1_04_en}
  \end{subfigure}
  \begin{subfigure}[t]{0.49\linewidth}
    \centering
    \includegraphics[width=\linewidth]{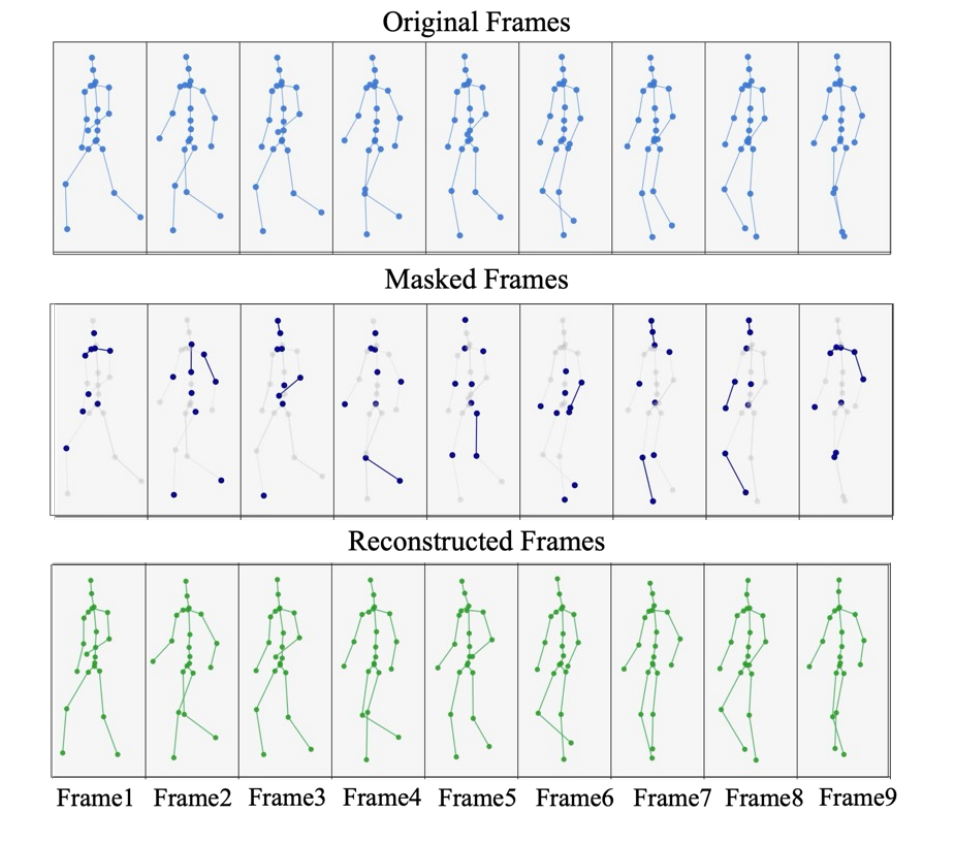}
    \caption{$r_{\mathrm{val}}=0.6$}
    \label{fig:recon_ex1_06_en}
  \end{subfigure}

\caption{
Reconstruction example 1  (successful case) under two validation masking ratios ($r_{\mathrm{val}}=0.4$ and $0.6$).
}
\label{fig:reconstruct_example1_ratios_en}
\end{figure}

\begin{figure}[t]
  \centering
  \begin{subfigure}[t]{0.49\linewidth}
    \centering
    \includegraphics[width=\linewidth]{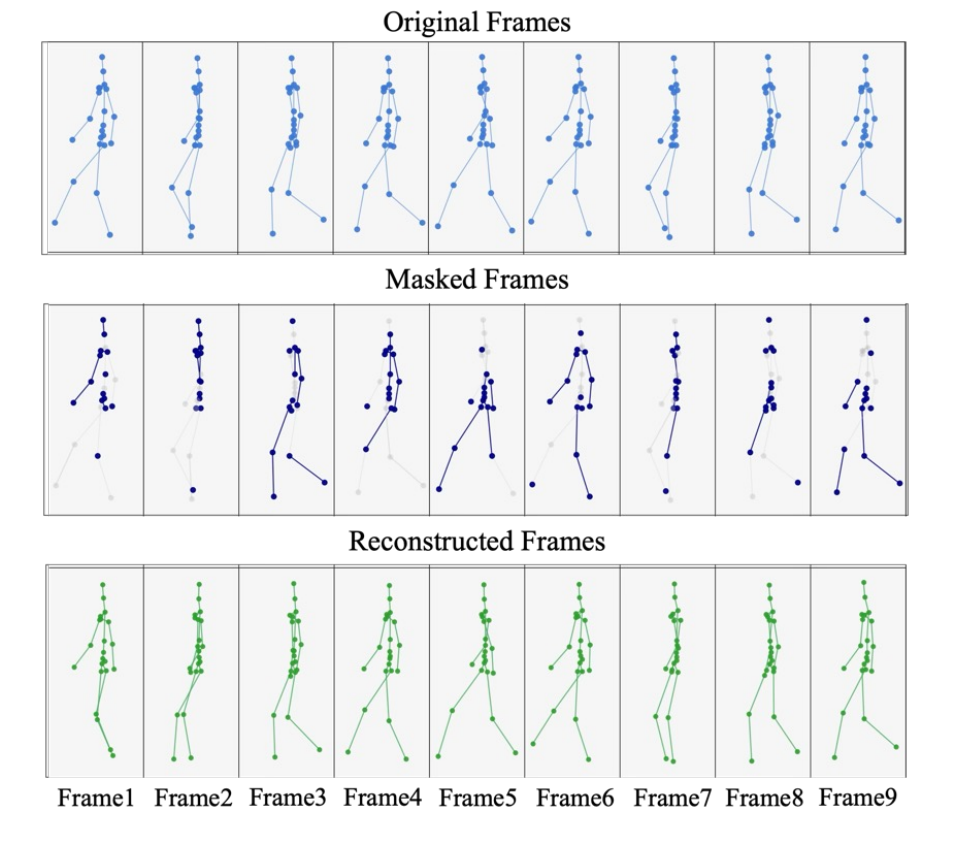}
    \caption{$r_{\mathrm{val}}=0.4$}
    \label{fig:recon_ex2_04_en}
  \end{subfigure}
  \begin{subfigure}[t]{0.49\linewidth}
    \centering
    \includegraphics[width=\linewidth]{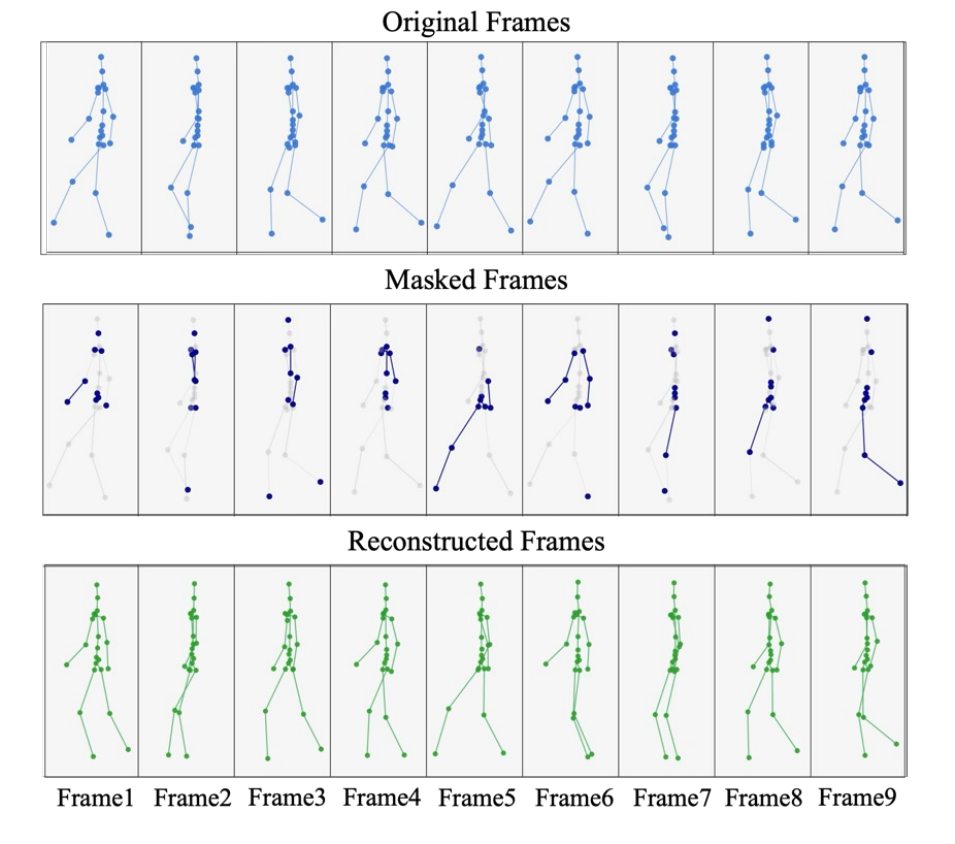}
    \caption{$r_{\mathrm{val}}=0.6$}
    \label{fig:recon_ex2_06_en}
  \end{subfigure}

\caption{
Reconstruction example 2  (successful case) under two validation masking ratios ($r_{\mathrm{val}}=0.4$ and $0.6$).
}
\label{fig:reconstruct_example2_ratios_en}
\end{figure}

\section{Additional Reconstruction Examples in Self-Supervised Skeleton Learning}
\label{app:qualitative_reconstruction}

\begin{figure}[t]
  \centering

  \begin{subfigure}[t]{0.49\linewidth}
    \centering
    \includegraphics[width=\linewidth]{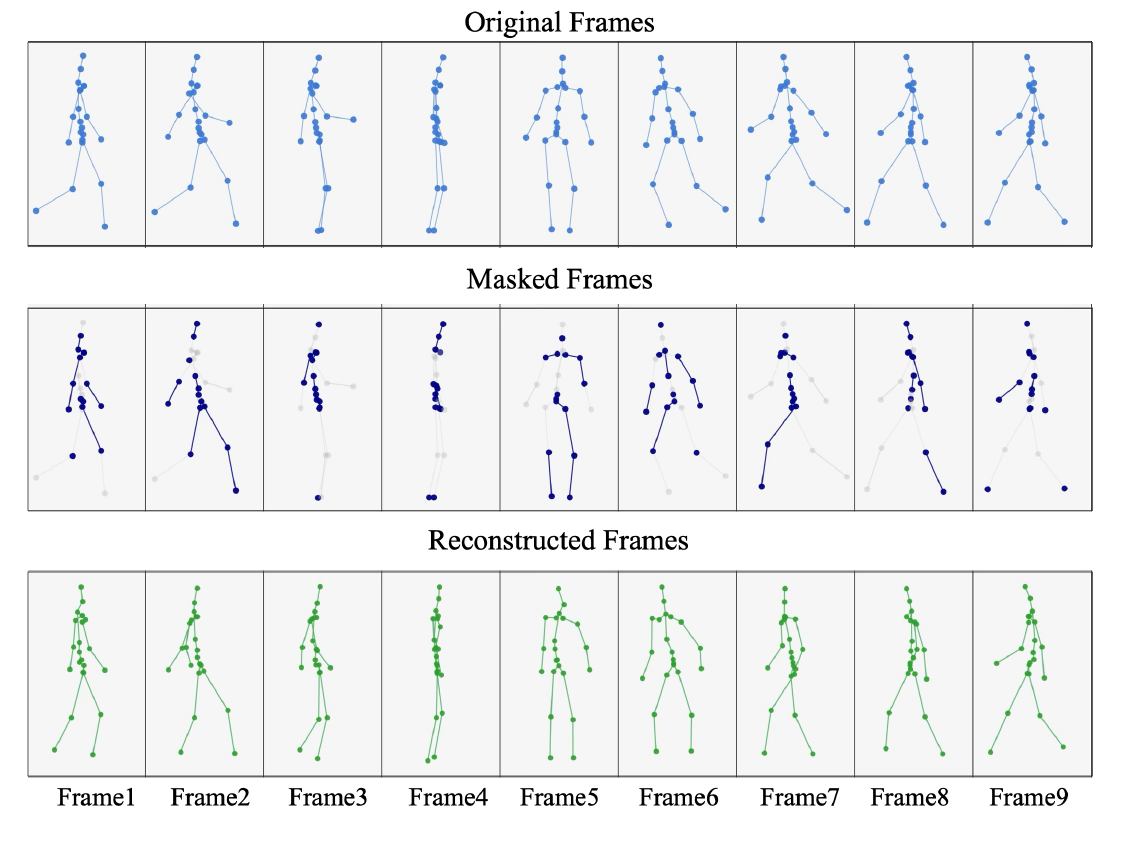}
    \caption{$r_{\mathrm{val}}=0.4$}
    \label{fig:recon_ex3_04_en}
  \end{subfigure}
  \begin{subfigure}[t]{0.49\linewidth}
    \centering
    \includegraphics[width=\linewidth]{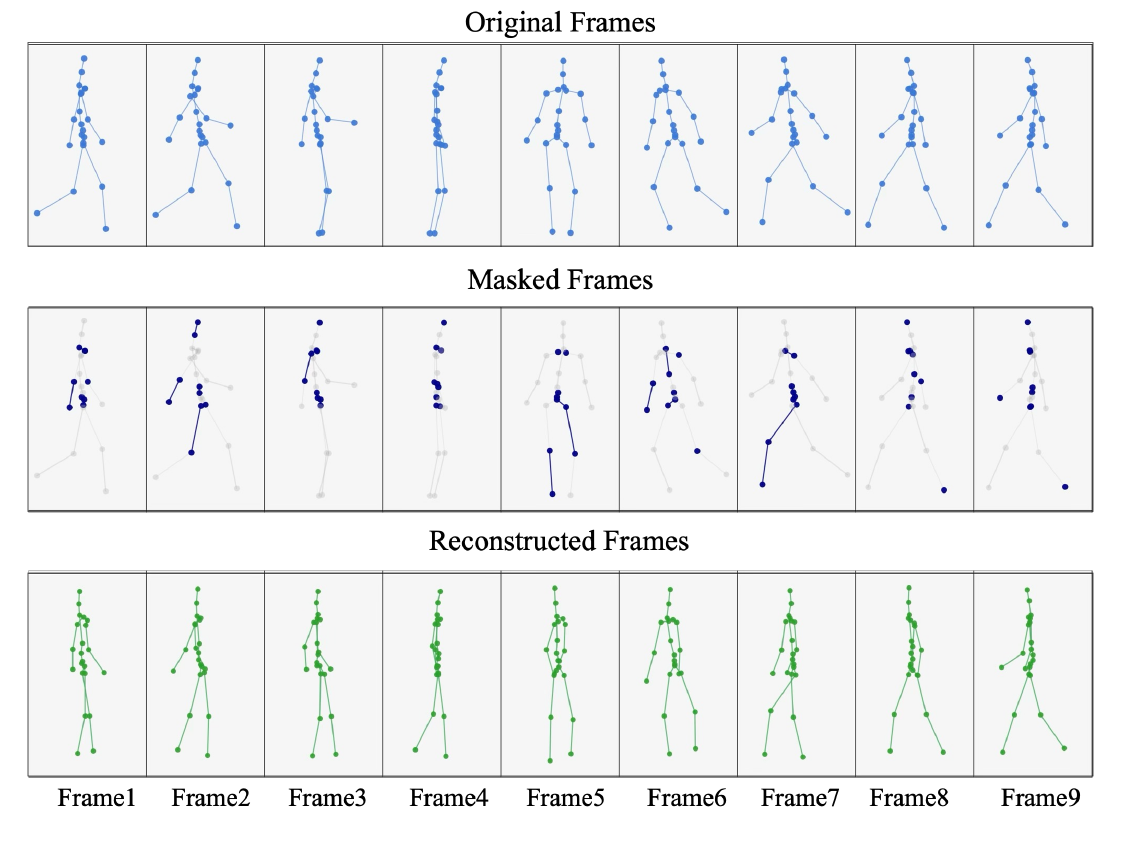}
    \caption{$r_{\mathrm{val}}=0.6$}
    \label{fig:recon_ex3_06_en}
  \end{subfigure}

\caption{
Reconstruction example 3  (challenging case) under two validation masking ratios ($r_{\mathrm{val}}=0.4$ and $0.6$).
}
\label{fig:reconstruct_example3_ratios_en}
\end{figure}

\begin{figure}[t]
  \centering

  \begin{subfigure}[t]{0.49\linewidth}
    \centering
    \includegraphics[width=\linewidth]{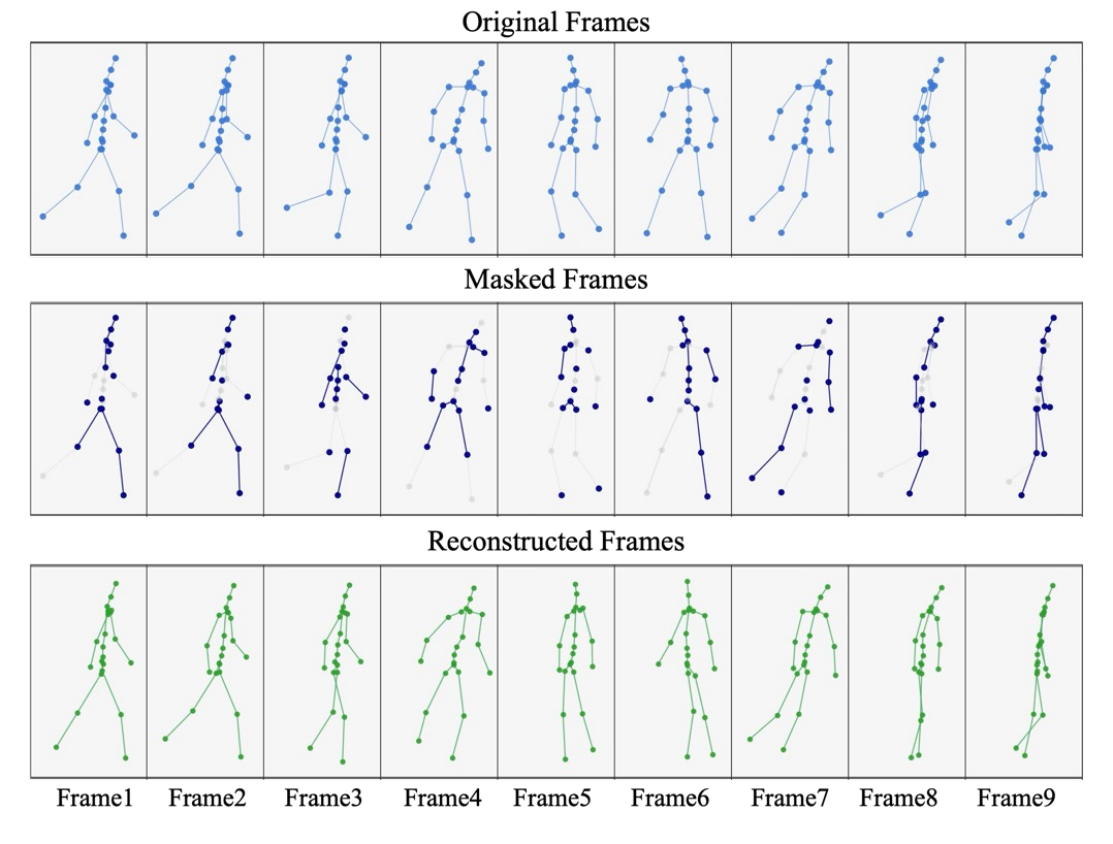}
    \caption{$r_{\mathrm{val}}=0.4$}
    \label{fig:recon_ex4_04_en}
  \end{subfigure}
  \begin{subfigure}[t]{0.49\linewidth}
    \centering
    \includegraphics[width=\linewidth]{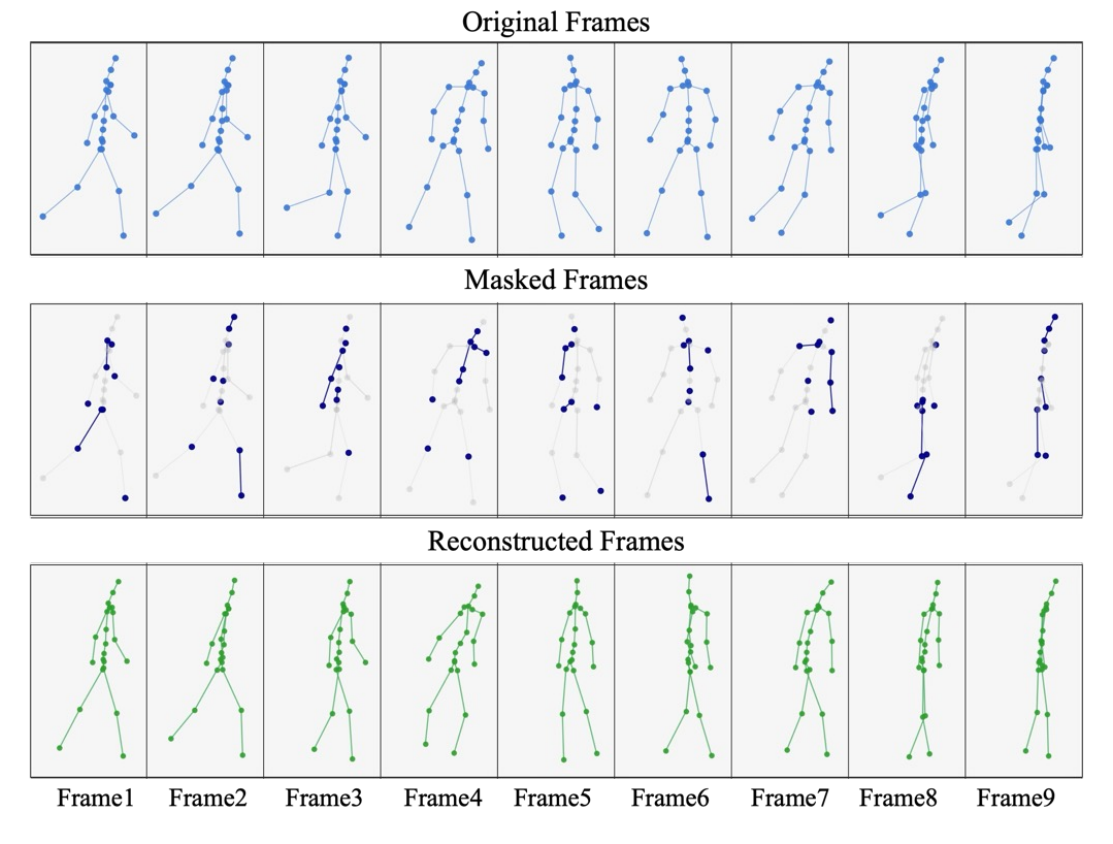}
    \caption{$r_{\mathrm{val}}=0.6$}
    \label{fig:recon_ex4_06_en}
  \end{subfigure}

\caption{
Reconstruction example 3  (challenging case) under two validation masking ratios ($r_{\mathrm{val}}=0.4$ and $0.6$).
}
\label{fig:reconstruct_example4_ratios_en}
\end{figure}

Figures~\ref{fig:reconstruct_example1_ratios_en}--\ref{fig:reconstruct_example4_ratios_en} show four qualitative examples of skeleton reconstruction under two validation masking ratios ($r_{\mathrm{val}}=0.4$ and $0.6$).
Examples 1 and 2 are representative successful cases, while Examples 3 and 4 are more challenging cases involving direction changes.

Examples 1 and 2 correspond to simple straight-motion patterns (with different traveling directions), and the model reconstructs them reliably at both masking ratios.
This indicates that the reconstruction model preserves the overall skeletal/motion structure well when the motion pattern is simple.

Examples 3 and 4 are harder cases with direction changes (one turn in Example 3 and two turns in Example 4).
At $r_{\mathrm{val}}=0.4$, the reconstructed skeletons can be locally noisy, but the turning behavior is still partially preserved.
At $r_{\mathrm{val}}=0.6$, the reconstructed poses become more averaged and body orientation is sometimes misestimated; however, the model often retains the coarse motion tendency.

Overall, these examples suggest that the reconstruction model is effective at preserving global motion cues under missingness, while fine-grained pose structure and orientation become more difficult to recover in highly masked, direction-changing sequences.

\section{Additional Analyses with Controlled Skeleton Reliance}
\label{sec:appendix_skeleton_contribution}

This appendix provides controlled analyses that complement the robustness results in the main text.
As discussed in Section~\ref{sec:robustness_eval}, robustness cannot be fairly compared from degradation trends alone, because the compared models may rely on skeleton cues to different degrees.
To reduce this confounding factor, we conduct controlled comparisons designed to match the skeleton encoder capacity and, as much as possible, the effective reliance on skeleton cues.

\subsection{Controlled Baseline: ST-GCN Replacement without Pretraining (Supervised Only)}
\label{app:controlled_baseline}

To enable a fairer robustness comparison, we introduce a controlled baseline that matches the skeleton encoder architecture of our method while removing self-supervised pretraining.
Specifically, we replace the original skeleton embedding module with the same ST-GCN architecture used in \emph{Ours}, but train it from scratch using only downstream trajectory supervision (i.e., without self-supervised pretraining).
We refer to this model as \emph{Baseline (ST-GCN)}.

This controlled baseline serves two purposes.
First, it reduces the architectural gap between the standard baseline and \emph{Ours}.
Second, it provides a comparison target with similar skeleton-encoder capacity, allowing us to better isolate the effect of self-supervised pretraining on robustness.

As shown in Table~\ref{tab:skeleton_ablation_controlled}, \emph{Baseline (ST-GCN)} and \emph{Ours} exhibit comparable performance drops in the skeleton-disabling test, indicating a similar level of skeleton contribution.
This makes the robustness comparison in the next subsection more interpretable.

\emph{Baseline (ST-GCN)} achieves a clean-input accuracy of $0.895/1.810$ (ADE/FDE), suggesting that upgrading the skeleton encoder alone encourages the predictor to exploit motion cues more effectively.
However, as shown next, this increased reliance on skeleton information can amplify sensitivity to missing joints.
We report its behavior under random joint masking in Appendix~\ref{app:controlled_robustness}.

\begin{table}[t]
\centering
\caption{
Controlled skeleton ablation comparison (clean input; mask rate $0.0$).
We compare our pretrained encoder against a same-parameter ST-GCN replacement trained \emph{without} pretraining.
This provides a reference with similar skeleton contribution ($\Delta$), isolating the effect of pretraining.
}
\label{tab:skeleton_ablation_controlled}
\setlength{\tabcolsep}{6pt}
\renewcommand{\arraystretch}{1.15}

\resizebox{\linewidth}{!}{
\begin{tabular}{lcccc}
\hline
\textbf{Model}
& \textbf{Clean} & \textbf{Skeleton disabled} & $\mathbf{\Delta}$\textbf{ADE} & $\mathbf{\Delta}$\textbf{FDE} \\
\hline
Baseline (Standard)
& 0.920 / 1.884
& 1.563 / 3.186
& +0.643 & +1.301 \\
Baseline (ST-GCN)
& 0.895 / 1.810
& 2.657 / 4.868
& +1.762
& +3.058 \\

Ours
& 0.898 / 1.832
& 2.389 / 4.789
& +1.491
& +2.957 \\
\hline
\end{tabular}
}
\end{table}
\begin{table}[t]
\centering
\caption{
Performance under random joint masking.
Each entry reports ADE / FDE. Bold indicates the best result at each mask rate.
}
\label{tab:joint_masking_3methods}
\setlength{\tabcolsep}{6pt}
\renewcommand{\arraystretch}{1.15}

\resizebox{\linewidth}{!}{
\begin{tabular}{lcccc}
\hline
\textbf{Method}
& \textbf{Clean (0.0)} & \textbf{Low (0.2)} & \textbf{Moderate (0.4)} & \textbf{High (0.6)} \\
\hline
Baseline (Standard)
& 0.920 / 1.884
& 0.940 / 1.916
& 0.969 / 1.957
& 1.050 / 2.090\\

Baseline (ST-GCN)
& \textbf{0.895/1.810}
& 1.060 / 2.094
& 1.330 / 2.646
& 1.543 / 3.106 \\

Ours 
& 0.898 / 1.832
& \textbf{0.913 / 1.853}
& \textbf{0.934 / 1.884}
& \textbf{1.017 / 2.058} \\
\hline
\end{tabular}
}
\end{table}

\subsection{Controlled Robustness under Random Joint Masking}
\label{app:controlled_robustness}
We evaluate robustness under random joint masking for three models:
(i) Baseline (standard), (ii) Baseline (ST-GCN; supervised only), and (iii) Ours (pretrained).
This controlled comparison helps disentangle robustness gains from differences in architectural capacity and skeleton dependence.

A key observation is that Baseline (ST-GCN) is markedly less robust than Baseline (standard).
This trend is consistent with the skeleton-contribution analysis in Section~\ref{sec:robustness_eval}:
because Baseline (ST-GCN) relies more on skeleton cues, missing joints more severely disrupt its predictions.
In other words, stronger skeleton utilization can improve clean-input performance while amplifying brittleness under corruption.

Importantly, despite exhibiting similarly high skeleton contribution, \emph{Ours} consistently outperforms Baseline (ST-GCN) under low-to-moderate masking, demonstrating that self-supervised pretraining yields more missing-tolerant representations beyond simply replacing the encoder.
Moreover, compared to Baseline (standard), \emph{Ours} remains superior up to $0.6$ masking, highlighting robustness gains in the practically relevant low-to-medium missingness regime.

We also note that the error increase of \emph{Ours} becomes more noticeable between $0.4$ and $0.6$ masking.
Beyond high missingness (e.g., $>0.6$), errors rapidly increase for all models, reflecting the ill-posed nature of trajectory prediction with extremely limited body observations.

\subsection{Controlled Comparison: Coordinate Completion vs.\ Latent Representation}
\label{app:coord_vs_latent}

The robustness ordering can change under severe missingness, partly because the results become sensitive to how strongly each model relies on skeleton cues.
To make the comparison more interpretable, we construct a controlled setting that keeps the downstream trajectory predictor architecture identical and isolates \emph{how} missing skeletons are handled:
(i) completing missing joints as \emph{coordinates} and feeding the completed skeleton to the predictor, versus
(ii) absorbing missingness through a pretrained encoder and feeding \emph{latent features} (without using reconstructed coordinates).

Specifically, all compared methods use the same downstream trajectory predictor (including the same in-model ST-GCN skeleton branch in architecture), and differ only in how skeleton information is injected:

\begin{itemize}
\item \textbf{Baseline (ST-GCN) + Reconstruction}: the downstream predictor is exactly the same as Baseline (ST-GCN).
In addition, we use a skeleton reconstruction model (\ref{subsec:self-supervised}) to complete missing joints.
At test time, missing joints are first reconstructed.

\item \textbf{Ours}: the downstream predictor architecture is again the same, but we initialize the in-model ST-GCN skeleton encoder with the self-supervised pretrained encoder and keep it \emph{frozen}.
The predictor receives the encoder's \emph{latent representations} as skeleton cues, and reconstructed joint coordinates are never passed downstream.
\end{itemize}

\begin{table}[t]
\centering
\caption{
Performance under random joint masking.
Each entry reports ADE / FDE. Bold indicates the best result at each mask rate.
The coordinate-completion baseline reconstructs missing joints and feeds completed \emph{coordinates} to the predictor.
}
\label{tab:joint_masking_4methods}
\setlength{\tabcolsep}{6pt}
\renewcommand{\arraystretch}{1.15}

\resizebox{\linewidth}{!}{
\begin{tabular}{lcccc}
\hline
\textbf{Method}
& \textbf{Clean (0.0)} & \textbf{Low (0.2)} & \textbf{Moderate (0.4)} & \textbf{High (0.6)} \\
\hline

Baseline (ST-GCN)
& \textbf{0.895/1.810}
& 1.060 / 2.094
& 1.330 / 2.646
& 1.543 / 3.106 \\

Baseline (ST-GCN) + Recon
& \textbf{0.895 / 1.810}
& \textbf{0.902 / 1.820}
& 0.969 / 1.952
& 1.357 / 2.781 \\

Ours
& 0.898 / 1.832
&0.913 / 1.853
& \textbf{0.934/1.884}
& \textbf{1.017 / 2.058} \\
\hline
\end{tabular}
}
\end{table}

Table~\ref{tab:joint_masking_4methods} provides a controlled comparison to disentangle whether skeleton cues help from how they should be injected into the predictor. In particular, the coordinate-completion baseline (Baseline (ST-GCN) + Recon) explicitly reconstructs missing joints and feeds the completed coordinates to the downstream trajectory predictor, whereas Ours never passes reconstructed coordinates and instead injects high-dimensional skeleton representations learned via self-supervised pretraining.

Under clean inputs, the gap between Baseline (ST-GCN) and Ours is very small. This indicates that our robustness-oriented pretraining can trade a negligible amount of clean-condition optimality for improved stability under partial observability, i.e., a minor loss of “clean-specialized” information in exchange for missing-tolerant representations.

As the missingness increases, the difference between coordinate-level completion and representation-level integration becomes more pronounced. While coordinate completion can yield gains at low missingness, its performance degrades substantially at higher mask rates. This trend is consistent with the fact that reconstruction errors inevitably grow with missingness and, when reconstructed coordinates are directly fed to the predictor, these errors are propagated and potentially amplified by the downstream model. In contrast, Ours exhibits a much milder degradation and achieves the best results at moderate-to-high missingness. These results support our design choice: incorporating skeleton information through robust latent representations is preferable to injecting reconstructed coordinates, because representation-level integration mitigates the direct transmission of reconstruction noise to trajectory prediction.
Overall, the table suggests that, as skeleton cues are leveraged more aggressively in trajectory prediction, avoiding coordinate-level error propagation becomes increasingly important; consequently, learning and using missing-tolerant high-level representations is a more reliable strategy than coordinate completion under severe joint missingness.

\section{Reconstruction-Based Completion is Complementary to the Proposed Method}
\label{app:ours_plus_recon}
We further examine whether reconstruction-based coordinate completion is an alternative to our method or a complementary component.
To this end, we apply the same reconstruction frontend used in \emph{Baseline (Reconstruction)} also to \emph{Ours}, and evaluate the combined model (\emph{Ours + Reconstruction}) under random joint masking.

Table~\ref{tab:ours_plus_recon} shows that reconstruction and the proposed pretrained representation are not mutually exclusive.
Compared with \emph{Ours}, \emph{Ours + Reconstruction} consistently improves ADE/FDE across all evaluated masking rates, with especially clear gains in the low-to-moderate masking regime.
This indicates that coordinate-level completion and representation-level robustness address different aspects of missingness and can be combined effectively.

Compared with \emph{Baseline (Reconstruction)}, \emph{Ours + Reconstruction} achieves lower errors at all masking rates in this setting, suggesting that the gain does not come only from reconstruction itself, but also from the missing-tolerant latent representation learned by our self-supervised pretraining.
These results support our claim that the proposed method is compatible with upstream pose/skeleton restoration modules, rather than being a competing alternative.

\begin{table}[t]
\centering
\caption{
Compatibility of reconstruction-based coordinate completion with the proposed method under random joint masking.
Each entry reports ADE / FDE. Lower is better.
}
\label{tab:ours_plus_recon}
\setlength{\tabcolsep}{6pt}
\renewcommand{\arraystretch}{1.12}
\resizebox{\linewidth}{!}{
\begin{tabular}{lcccc}
\hline
\textbf{Method}
& \textbf{Clean (0.0)} & \textbf{Low (0.2)} & \textbf{Moderate (0.4)} & \textbf{High (0.6)} \\
\hline
Baseline (Reconstruction)
& 0.920 / 1.884
& 0.927 / 1.892
& 0.940 / 1.907
& 0.973 / 1.939 \\

Ours
& \textbf{0.898} / \textbf{1.832}
& 0.913 / 1.853
& 0.934 / 1.884
& 1.017 / 2.058 \\

Ours + Reconstruction
& \textbf{0.898} / \textbf{1.832}
& \textbf{0.903} / \textbf{1.838}
& \textbf{0.910} / \textbf{1.849}
& \textbf{0.933} / \textbf{1.878} \\
\hline
\end{tabular}
}
\end{table}

\section{Qualitative Case Study on Prediction Updates Under Clean Skeleton Inputs}
\label{sec:appendix_pose_update_clean}

In this appendix, we provide a qualitative case study under clean skeleton inputs to support that the proposed method extracts and utilizes more informative skeleton-based motion features than \emph{Baseline (Standard)}.
We focus on a scene in which the future turning behavior is not yet evident at an earlier observation step, but becomes more predictable two frames later as body orientation and pose configuration evolve.

\begin{figure}[t]
  \centering
  \begin{subfigure}[t]{0.49\linewidth}
    \centering
    \includegraphics[width=\linewidth]{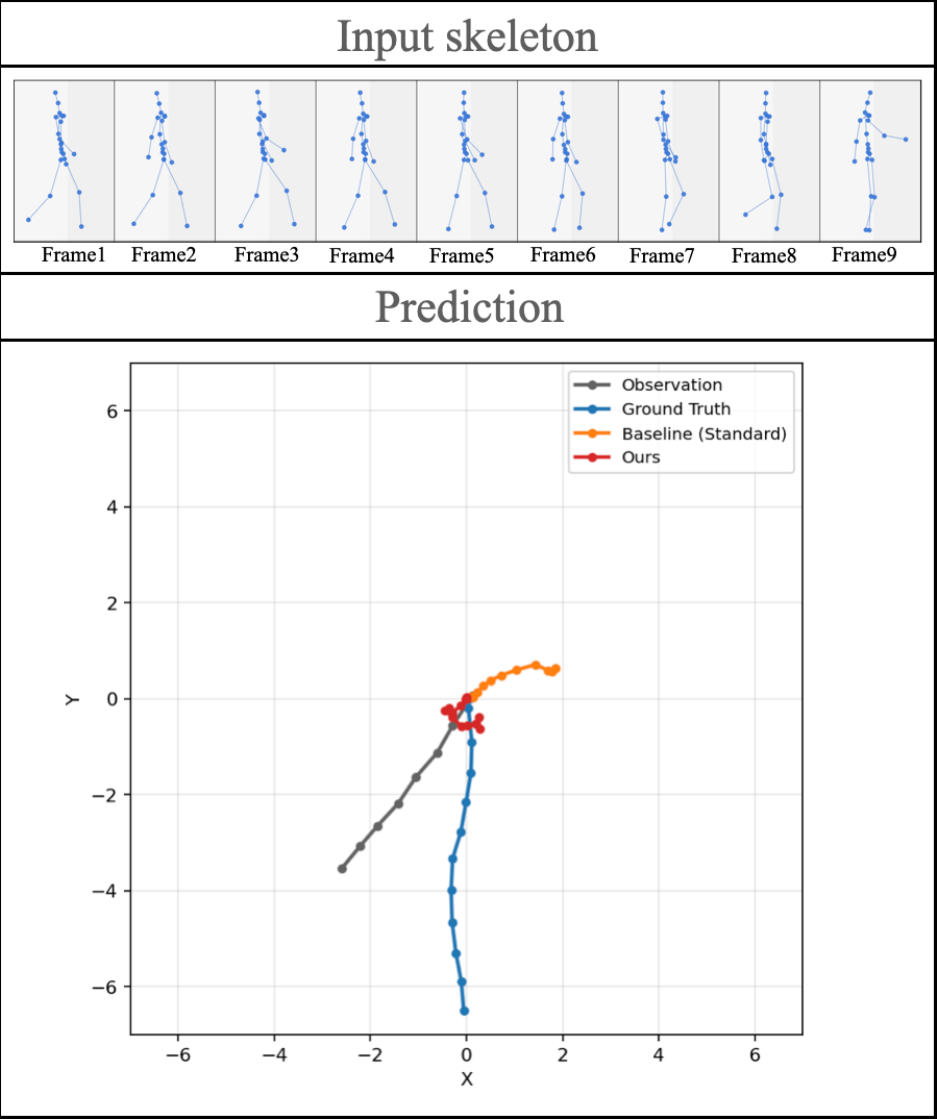}
    \caption{Earlier observation step}
    \label{fig:pose_update_case_early}
  \end{subfigure}\hfill
  \begin{subfigure}[t]{0.49\linewidth}
    \centering
    \includegraphics[width=\linewidth]{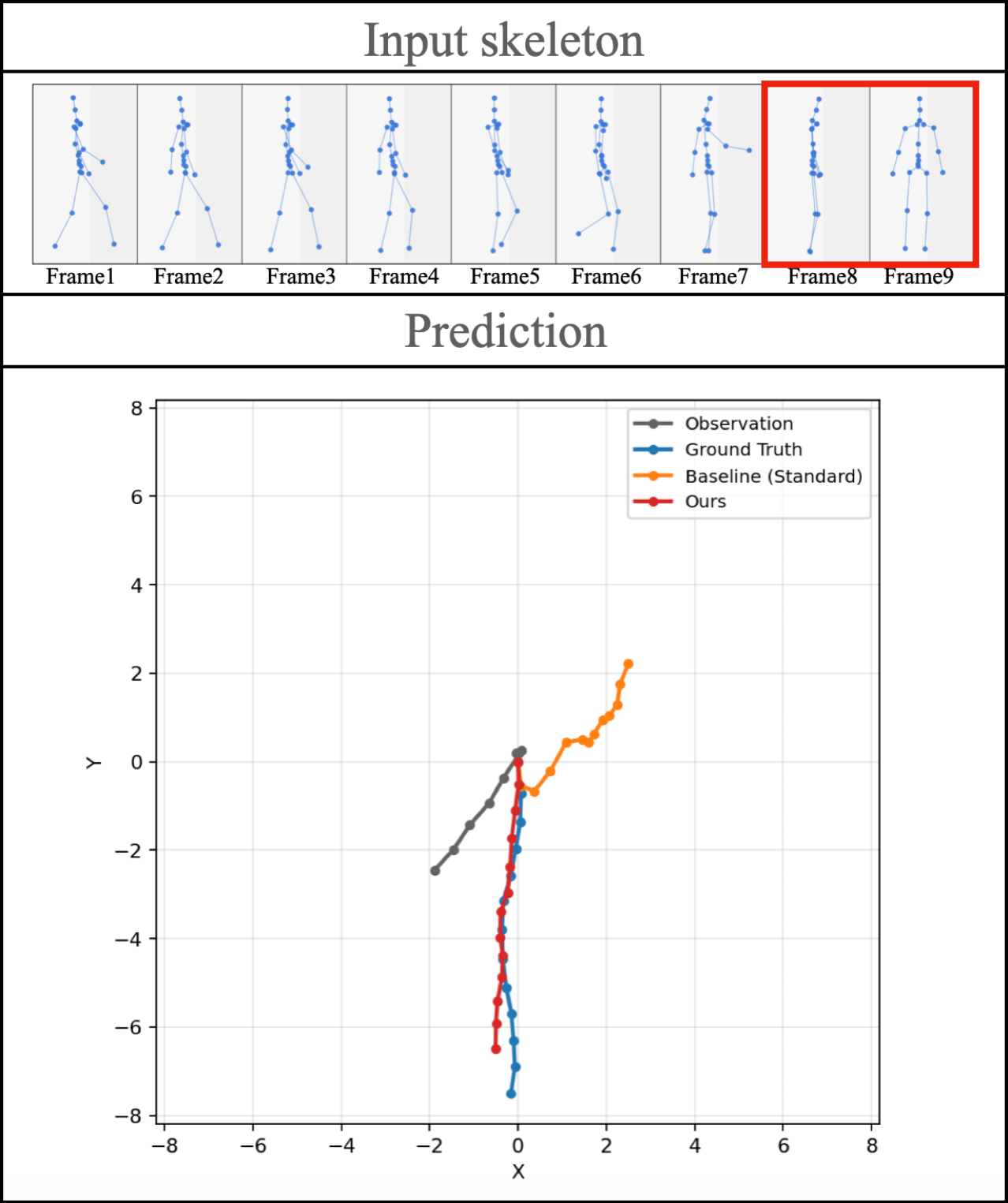}
    \caption{Two frames later}
    \label{fig:pose_update_case_late}
  \end{subfigure}
  \caption{
  Qualitative case study on prediction updates under clean skeleton inputs.
  The same scene is shown at two observation steps separated by two additional observed frames.
  In each panel, the top row shows the input skeleton sequence and the bottom row shows the predicted trajectories by \emph{Baseline (Standard)} and \emph{Ours} with the ground truth.
  In the right panel, the two newly observed frames are highlighted with red boxes.
  The later observation reveals a clearer body-orientation cue for the upcoming turn.
  While both methods are inaccurate at the earlier step, \emph{Ours} updates its prediction toward the curved ground-truth trajectory after the additional observations, whereas \emph{Baseline (Standard)} remains less responsive.
  }
  \label{fig:appendix_pose_update_clean}
\end{figure}

Figure~\ref{fig:appendix_pose_update_clean} shows the same scene at two observation steps separated by two additional observed frames.
Each panel contains the input skeleton sequence (top) and the predicted trajectories (bottom).
In the later observation step, the two newly observed frames are highlighted with red boxes.
Notably, the final observed frame reveals a clearer change in body orientation, which provides an important cue for the upcoming direction change.

At the earlier observation step, both \emph{Baseline (Standard)} and \emph{Ours} are inaccurate, indicating that the future turning tendency is still ambiguous from the currently available motion evidence.
Two frames later, however, the pose becomes more informative.
At this point, \emph{Ours} updates its prediction more clearly toward the curved ground-truth trajectory, while \emph{Baseline (Standard)} remains less responsive and tends to stay closer to a straight prediction.

These examples suggest that the proposed method is more effective at extracting and utilizing motion-relevant skeleton cues when such cues emerge over time.
Importantly, the claim here is not that the proposed method always predicts correctly from limited observations.
Rather, the key qualitative evidence is whether the model updates its prediction when newly observed pose information becomes informative.
From this viewpoint, the proposed method demonstrates stronger skeleton feature extraction and usage than \emph{Baseline (Standard)}.

\section{Qualitative Case Study on Curved Trajectories}
\label{sec:appendix_curve_case}

In this appendix, we provide qualitative comparisons on curved-trajectory scenes to clarify when skeletal cues are essential and how the proposed method behaves under partial joint missingness.
We compare methods with relatively low skeleton reliance
(\emph{Baseline (Standard)}, \emph{Baseline (Reconstruction)}, and \emph{Baseline (Corruption-trained)})
and higher skeleton reliance (\emph{Baseline (STGCN)} and \emph{Ours}) in a unified figure.

\begin{figure}[t]
  \centering

  \begin{subfigure}[t]{0.49\linewidth}
    \centering
    \includegraphics[width=\linewidth]{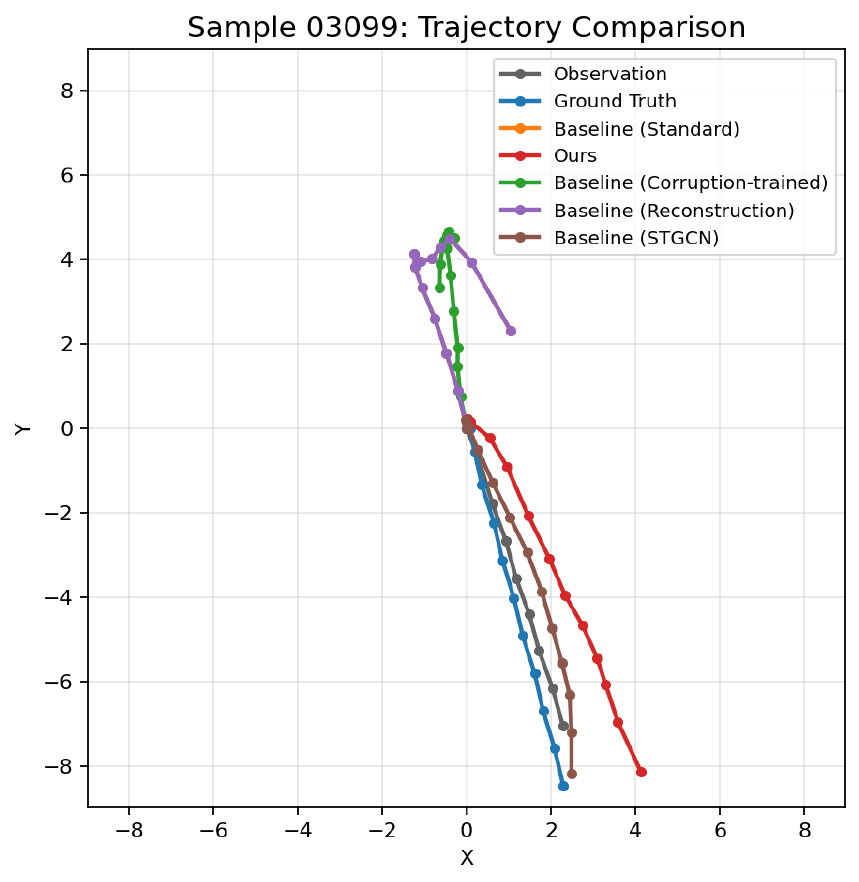}
    \caption{Case 1 (Clean)}
    \label{fig:curve_case1_clean}
  \end{subfigure}\hfill
  \begin{subfigure}[t]{0.49\linewidth}
    \centering
    \includegraphics[width=\linewidth]{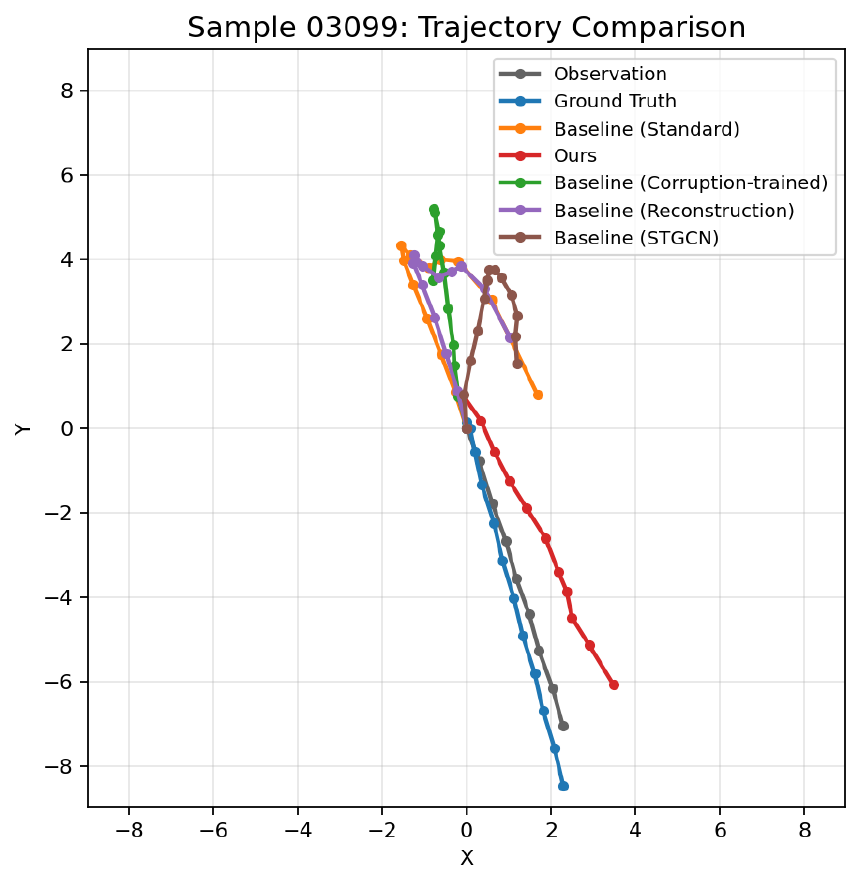}
    \caption{Case 1 (Masked, $r=0.4$)}
    \label{fig:curve_case1_mask04}
  \end{subfigure}

  \vspace{0.6em}

  \begin{subfigure}[t]{0.49\linewidth}
    \centering
    \includegraphics[width=\linewidth]{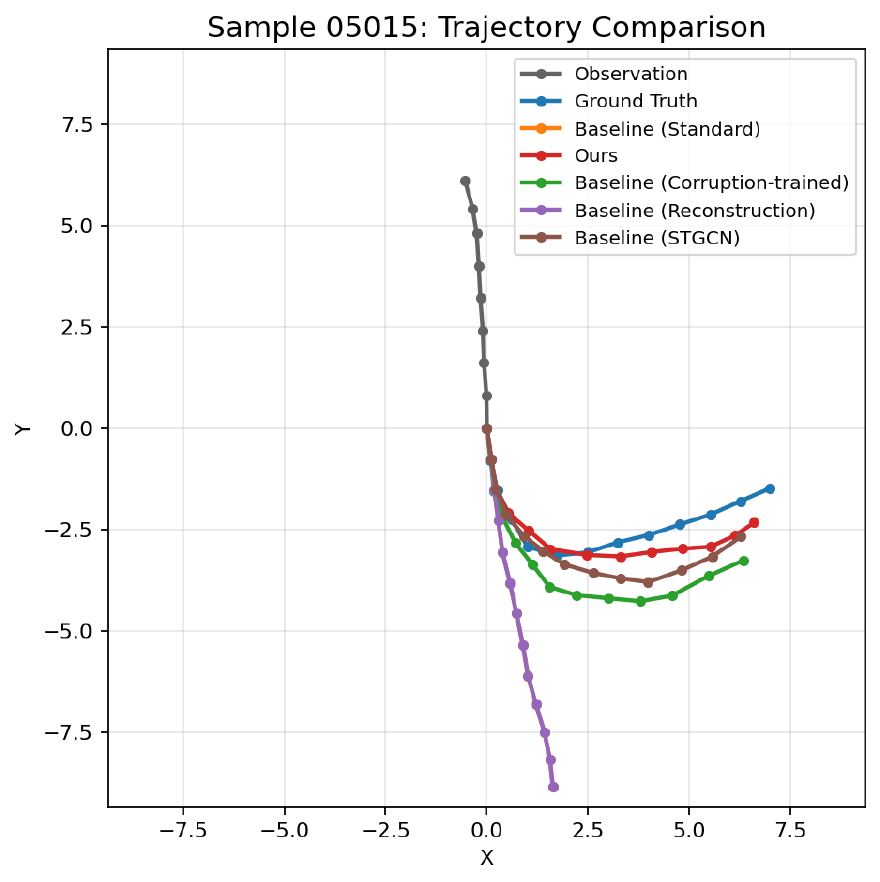}
    \caption{Case 2 (Clean)}
    \label{fig:curve_case2_clean}
  \end{subfigure}\hfill
  \begin{subfigure}[t]{0.49\linewidth}
    \centering
    \includegraphics[width=\linewidth]{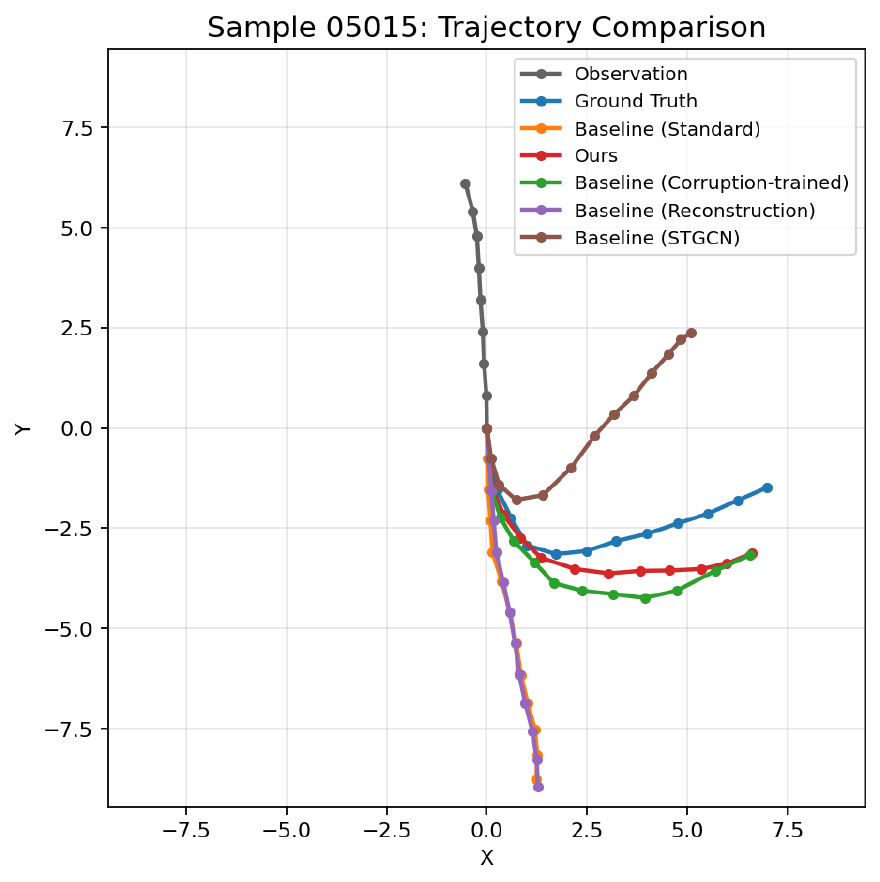}
    \caption{Case 2 (Masked, $r=0.4$)}
    \label{fig:curve_case2_mask04}
  \end{subfigure}

  \caption{
  Qualitative comparison on curved-trajectory scenes under clean and partially missing skeleton inputs.
Methods that effectively exploit skeletal cues better capture turning motion under clean inputs, whereas methods with low skeleton reliance tend to produce overly straight predictions.
Under random joint masking ($r=0.4$), the proposed method better preserves the turning tendency than the baselines, supporting improved robustness through more stable skeleton-based motion cues.
  }
  \label{fig:appendix_curve_cases}
\end{figure}

As shown in Fig.~\ref{fig:appendix_curve_cases}, under clean skeleton input, the low-reliance methods often fail to capture path curvature and tend to produce overly straight predictions.
In contrast, the higher-reliance methods (\emph{Baseline (STGCN)} and \emph{Ours}) better follow the turning tendency in both examples.
This suggests that skeletal cues are crucial for curved-trajectory prediction, where body orientation and motion tendency provide important signals for anticipating future direction changes.

A clearer difference appears under random joint masking with $r=0.4$ (Fig.~\ref{fig:appendix_curve_cases}).
Although \emph{Baseline (STGCN)} can capture the turn under clean input, it often fails to maintain the turning tendency once skeleton observations become incomplete.
By contrast, \emph{Ours} remains closer to the curved ground-truth trajectory and preserves a prediction pattern similar to the clean condition.

These examples support the interpretation that the proposed method does not gain robustness by reducing skeleton reliance.
Instead, it improves robustness by preserving the usefulness of skeleton-based motion representations under partial observations.
This qualitatively supports that the proposed method mitigates the accuracy--robustness trade-off in skeleton-aided trajectory prediction.

\section{Model Size and Inference Time}
\label{app:model_size_speed}

To assess the computational cost of the proposed design and its practical variants, we compare the number of parameters and the inference time per sample across all methods.
Table~\ref{tab:model_size_speed} reports both model size and inference time in \emph{ms/sample}.
Overall, all methods are in a similar computational range.
Compared with \emph{Baseline (Standard, Corruption-trained)}, \emph{Ours} (and \emph{Baseline (ST-GCN)}) increases the number of parameters from 3.19M to 3.70M, while the inference time increases only slightly from 1.942 to 2.099\,ms/sample.
The reconstruction-based variants require a few more parameters and slightly longer inference time (\emph{Baseline (Reconstruction)}: 3.72M, 2.132\,ms/sample; \emph{Ours + Recon}: 4.23M, 2.273\,ms/sample), but the overall differences remain modest.

These results indicate that the proposed method and its related variants introduce only limited computational overhead.
In particular, the proposed method improves robustness while maintaining practical inference efficiency.

\begin{table}[t]
\centering
\caption{
Comparison of the number of parameters and inference time (ms/sample).
}
\label{tab:model_size_speed}
\resizebox{\linewidth}{!}{
\begin{tabular}{lcc}
\toprule
\textbf{Method} & \textbf{Params} & \textbf{Inference Time} \\
\midrule
Baseline (Standard/Corruption-trained) & 3,188,546 & 1.942 \\
Ours/Baseline (ST-GCN) & 3,703,198 & 2.099 \\
Baseline (Reconstruction) & 3,720,097 & 2.132 \\
Ours + Recon & 4,234,749 & 2.273 \\
\bottomrule
\end{tabular}}
\end{table}

\end{document}